\documentclass[letterpaper]{article} 
\usepackage{aaai2026}  
\usepackage{times}  
\usepackage{helvet}  
\usepackage{courier}  
\usepackage[hyphens]{url}  
\usepackage{graphicx} 
\usepackage{fancyvrb}
\usepackage{natbib}  
\usepackage{caption} 
\usepackage{algorithm}
\usepackage{algorithmic}
\usepackage{amsmath}
\usepackage{enumitem}
\usepackage[T1]{fontenc}

\usepackage{newfloat}
\usepackage{listings}
\DeclareCaptionStyle{ruled}{labelfont=normalfont,labelsep=colon,strut=off} 
\floatstyle{ruled}
\newfloat{listing}{tb}{lst}{}
\floatname{listing}{Listing}
\title{Can LLMs Generate High-Quality Task-Specific Conversations?}
\author{
    Shengqi Li,  Amarnath Gupta
}
\affiliations{
    \textsuperscript{\rm 1}San Diego Supercomputer Center\\
    University of California San Diego\\

    9500 Gilman Drive, MC 0505\\
    La Jolla, CA 92093 USA\\
    \{shl142\}@ucsd.edu, 
   \{a1gupta\}@ucsd.edu
}

\begin{document}

\maketitle

\begin{abstract}
This paper introduces a parameterization framework for controlling conversation quality in large language models. We explore nine key parameters across six dimensions that enable precise specification of dialogue properties. Through experiments with state-of-the-art LLMs, we demonstrate that parameter-based control produces statistically significant differences in generated conversation properties. Our approach addresses challenges in conversation generation, including topic coherence, knowledge progression, character consistency, and control granularity. The framework provides a standardized method for conversation quality control with applications in education, therapy, customer service, and entertainment. Future work will focus on implementing additional parameters through architectural modifications and developing benchmark datasets for evaluation.
\end{abstract}


\section{Introduction}
\label{sec:intro}
Generative AI represents a transformative class of artificial intelligence systems capable of autonomously producing diverse content based on patterns learned from large-scale data and guided by user prompts. These models can generate coherent and contextually relevant text~\cite{openai2024gpt4technicalreport}, synthesize photorealistic images~\cite{rombach2022high}, compose original music~\cite{copet2023simple}, produce functional source code~\cite{chen2021evaluatinglargelanguagemodels}, and design 3D models and environments~\cite{poole2022dreamfusiontextto3dusing2d}. Their generative capacity extends beyond creative tasks, with applications in scientific domains, such as predicting protein structures with atomic accuracy~\cite{jumper2021highly} and assisting in the formulation of mathematical proofs~\cite{drori2022neural}. This versatility has made generative AI a central technology in both creative industries and scientific research. 

We believe that in future, we will see the rise of \textbf{parametrically controlled LLMs} that are tuned to perform specific complex tasks, and will allow for finer-grained control of their behavior through a set of parameters instead of relying solely on natural language instructions. In this paper, we investigate a specific case to illustrate the point - the example task is that of generating \textit{realistic end-to-end multi-turn conversation}s using large language models (LLMs) as a means to simulate dialogue episodes \textit{in a given thematic area}. These simulated conversations would serve as structured training material that can improve downstream conversational AI applications, particularly in settings where data is scarce, human annotation is costly, or domain specificity is high. Generating whole conversations, rather than isolated responses, enables the development of systems that better capture context, discourse coherence, and speaker intent over extended interactions~\cite{zhang2020dialogptlargescalegenerativepretraining,roller2021recipes}.

Such simulators are increasingly important in real-world research areas ranging from healthcare and education to business advising and civic services. For instance, in training AI systems to support low-resource users—such as entrepreneurs seeking regulatory or startup guidance—few-shot or domain-specific conversations are essential, but often unavailable~\cite{tan2023chatdoctor}. Simulated dialogues can bridge this gap by providing varied, context-rich conversations tailored to user needs and grounded in realistic scenarios~\cite{huang2020challenges},
and allow researchers to probe system behaviors in a controlled manner—enabling stress testing for safety, bias, and human-centric factors ~\cite{bender2021dangers} like cognitive overload. 
Realistic multi-turn conversation generation is not just a technical convenience—it is emerging as a core methodology for training and evaluating next-generation dialogue systems.

In this paper, we introduce a parameterization framework for LLM-based conversation generation. Unlike unstructured prompting approaches, this parameterization enables precise specification of conversation properties that can be systematically varied, measured, and optimized. This approach builds upon prior work in controlled text generation~\cite{keskar2019ctrlconditionaltransformerlanguage, dathathri2019plug, khalifa2021distributional} but extends these techniques specifically for multi-turn dialogue contexts with novel parameter dimensions.

The need for parameterized conversation control is particularly acute in domains requiring high-quality simulated dialogues, such as training data generation for conversational AI systems~\cite{li2016diversitypromotingobjectivefunctionneural}, educational dialogue design~\cite{nye2014autotutor}, therapeutic conversation modeling~\cite{vaidyam2019chatbots}, and realistic character interactions in entertainment applications~\cite{shuster2022blenderbot3deployedconversational, urbanek2019learning}. Recent work by \cite{zheng2023judging} demonstrates that conversation quality assessment is multidimensional, yet current generation approaches lack explicit control over these dimensions. While current LLMs can generate plausible conversations, they face several challenges that our parameterization approach directly addresses:

\begin{enumerate}
\item \textbf{Challenges in conversation quality control}
\begin{itemize}[leftmargin=*]
\item \textbf{Structural Coherence}: LLMs demonstrate documented difficulties maintaining consistency across extended dialogues. Research by \cite{gao2018neural} confirms a deterioration in response quality as conversation history increases, while \cite{xu2022beyond} identifies specific challenges in entity tracking and resolution of coreference over multiple turns. More recent studies by \cite{dziri2022origin} quantify inconsistencies in model-generated dialogues, showing that even state-of-the-art models exhibit significant contradiction rates. Our framework addresses these issues through explicit parameters for narrative coherence, memory utilization, and contradiction detection, building on techniques from computational narratology~\cite{mani2012computational}.
    \item \textbf{Knowledge Progression}: Studies by \cite{kim2020designing} show that effective knowledge transfer in educational dialogues requires careful calibration of complexity progression. Our parameters for explanation progression, conceptual density, and learning framework provide fine-grained control over knowledge transfer dynamics, drawing on established pedagogical frameworks~\cite{bloom1956taxonomy, anderson2001taxonomy} and cognitive load theory~\cite{sweller2011cognitive}.
    
    \item \textbf{Character Consistency}: Current approaches struggle to maintain consistent character voices and knowledge states throughout extended conversations. \cite{li2016persona} and \cite{zhang2018personalizing} demonstrate that explicit persona modeling improves response consistency, but challenges persist in maintaining these personas across turns. Our parameterization includes explicit controls for character consistency, knowledge asymmetry, and backstory depth to address these challenges, incorporating insights from computational models of personality~\cite{mairesse2007using} and literary character development~\cite{bamman2013learning, bamman2014bayesian}.
\end{itemize}

\item \textbf{Challenges in conversation generation methodology}
\begin{itemize}
    \item \textbf{Control Granularity}: Existing approaches typically offer coarse-grained control through natural language instructions, which can be ambiguous and inconsistently interpreted by models~\cite{mishra2022reframing, sanh2022multitask}. Recent work by \cite{min2022rethinking} shows significant variance in how models interpret the same natural language instructions. Our parameterization aim to provide control over conversation properties, similar to approaches in other generative domains such as text-to-image generation~\cite{nichol2022glidephotorealisticimagegeneration} and music synthesis~\cite{agostinelli2023musiclmgeneratingmusictext}.
    
    \item \textbf{Theoretical Grounding}: Current conversation generation approaches often lack connection to established theoretical frameworks in linguistics and dialogue management. \cite{larsson2000information} and \cite{traum2003information} provide formal models of tracking the state of dialogue that have not been fully used in the generation of neural conversations. Our parameter set establishes formal connections to speech act theory~\cite{searle1969speech, austin1975things}, information theory~\cite{shannon1948mathematical}, computational narratology~\cite{mani2012computational}, and dialogue management models~\cite{young2013pomdp, williams2016dialog}, creating a bridge between neural approaches and classical dialogue system theory.
    
    \item \textbf{Evaluation Framework}: \cite{deriu2020survey} identifies significant gaps in conversation evaluation methodologies, a finding echoed by \cite{mehri2020usrunsupervisedreferencefree}, who demonstrate poor correlation between automated metrics and human judgments of conversation quality. \cite{see2019makesgoodconversationcontrollable} further shows that human quality assessments depend on multiple dimensions that current automatic metrics do not capture comprehensively. Our parameterization approach enables systematic variation of conversation properties, facilitating controlled experiments to assess quality dimensions and potentially leading to more nuanced evaluation methodologies.
\end{itemize}
\end{enumerate}


The key contributions of this paper are:
\begin{enumerate}
\item A comprehensive taxonomy of 35 conversation parameters with 9 dominating factors organized into six dimensions that capture the essential aspects of high-quality conversations, extending prior work on dialogue quality factors~\cite{see2019makesgoodconversationcontrollable, mehri2020usrunsupervisedreferencefree}
\item Analysis of parameter necessity and sufficiency, identifying a core set of essential parameters while eliminating redundancies, informed by dimensionality reduction approaches to conversation modeling~\cite{larochelle2009exploring, lowe2018automaticturingtestlearning}
\item Formal theoretical connections between our parameters and established models in computational linguistics~\cite{jurafsky2000speech}, dialogue management~\cite{young2013pomdp}, and information theory~\cite{xu2020theory}, creating a bridge between neural approaches and classical dialogue system theory
\item Preliminary experimental validation demonstrating how modern LLMs can effectively implement a subset of these parameters through prompt conditioning, building on recent advances in controlled text generation~\cite{khalifa2021distributional, krause2020, Yang_2021}
\item A proposed research agenda for implementing the full parameter set through architectural modifications~\cite{hu2017toward, keskar2019ctrlconditionaltransformerlanguage}, developing efficient parameter encoding methods~\cite{li2021prefixtuningoptimizingcontinuousprompts, lester2021powerscaleparameterefficientprompt}, and creating benchmark datasets~\cite{welleck2019dialoguenaturallanguageinference, dziri2020evaluatingcoherencedialoguesystems}
\end{enumerate}
Our methodology combines computational approaches with insights from linguistics, psychology, and education. We evaluate our framework through a series of controlled experiments comparing conversations generated with systematically varied parameter settings. Results demonstrate statistically significant differences in generated conversation properties when parameter values are manipulated, confirming the effectiveness of our approach for a subset of parameters. For parameters that current LLMs struggle to implement reliably, we provide a detailed analysis of limitations and propose architectural modifications to address these challenges.

Our parameterization framework represents a significant step toward more controllable, higher-quality conversation generation with LLMs. By providing a standardized approach to conversation quality control, we aim to influence the theoretical understanding and practical capabilities of conversational AI systems. 

\section{Evaluation Tasks}
Here, we first introduce our evaluation tasks and explain the methods in Section 3. 
\paragraph{Topic Diversity}
The conversation needs a topic to start. After setting the topic area before the simulation, LLM will pick a subtopic based on the configured parameters to best suit the entrepreneur's background. In this task, we compare the distributions of topics mentioned by the simulator. 

\paragraph{Parameter Adherence}
To evaluate whether the conversation generated follows the given parameters, we evaluate the difference between the settled parameters vs. the inferred parameters given only the generated conversation. 

\paragraph{Topic Drift}
Natural dialogue often involves gradual topic transitions that can lead to substantial drift from the original subject matter, making thematic coherence throughout extended conversations a challenge. We measure the semantic distance between conversation segments to quantify how far the dialogue deviates from its initial topic focus. We calculate sentence embedding to compute cosine similarity scores between the opening conversational topic and subsequent dialogues, tracking the drift over turns. 

\paragraph{Character Properties Stability}
Consistent character portrayal across conversation turns is essential for believable dialogues, yet current LLMs often exhibit personality inconsistencies that undermine conversation quality. This evaluation measures character stability by analyzing linguistic markers, decision-making patterns, and domain expertise demonstrations throughout generated conversations. We measure deviations between the character's behavior in conversation versus their given background or parameters.

\paragraph{Entity Revisit Rate}
Effective conversations demonstrate sophisticated information management by strategically reintroducing previously mentioned entities, concepts, and topics, creating coherent narrative threads rather than generating unrelated information. We quantify how frequently and effectively the conversation simulator references earlier elements by tracking named entities and key concepts from earlier turns, then analyzing whether their subsequent appearances serve meaningful conversational purposes.

\section{Methods}

For this exploratory study, we selected nine parameters from the 35 that are the dominant factors of conversation quality, which are spread across the six dimensions.
\begin{itemize}
\item \textit{Turn}: The number of turns of the conversation.
\item \textit{Industry Context}: The initial field of this conversation.
\item \textit{Knowledge Gap Level}: The prior knowledge the entrepreneur has of the conversation's field. This is a method used in ~\cite{baskar2025guessingaskingapproachresolving} to measure the model's knowledge alignment with the entrepreneur. We define the gap as a 1-5 integer value, where 1 refers to an expert with a deep understanding of the domain, and 5 refers to a complete novice with minimal business knowledge about their ideas.
\item \textit{Smoothness Factor}: A grade A-F indicating conversation flow, with A referring to a perfectly flowing conversation with logical transitions, and F referring to a highly disjointed conversation with random topic jumping.
\item \textit{Focus Level}: A grade 1-5 indicating how focused the entrepreneur is on this conversation. 1 refers to free-flowing, wide-ranging conversation covering many aspects, and 5 refers to laser-focused on specific details of implementation.
\item \textit{Identity}: The initial setting of the entrepreneur's background, which is used by ~\cite{aher2023using} to simulate gender and racial diversity.
\item \textit{Technical Language Level}: A 0-1 float number indicating the level of technical language the entrepreneur is using in the conversation. Similar methods were used in ~\cite{scarlatos2025exploring} to trace knowledge levels in system-user conversation.
\item \textit{Formality Level}: A 0-1 float number indicating the formal phrase usage in the conversation.
\item \textit{Decision-Making Style}: The style of response the entrepreneur treats the system's response. It can be one of analytical, Intuitive, consultative, or impulsive.
\end{itemize}
The exact definition of other parameters used in the prompt, the precise definition of the value of each parameter, and examples can be found in the Appendix. 

\paragraph{Prompt Engineering}

The data set is created by constructing parameterized prompts that combine three key components: a base conversation generation prompt specifying the business advisory scenario, detailed parameter definitions for each dimension, and the specific parameter values for each conversation instance. For each experimental condition, we systematically vary the parameter values while maintaining consistent entrepreneur background profiles and industry contexts. The final prompt is fed to the target LLM to generate complete multi-turn conversations. The full prompt structure with an example implementation is in the Appendix. 

\paragraph{Model Selection}
We evaluate four state-of-the-art LLMs: \textit{Gemini-2.5-pro} ~\cite{comanici2025gemini25pushingfrontier}, \textit{Claude-3.7-sonnet} ~\cite{bai2022constitutionalaiharmlessnessai}, \textit{o3}, \textit{o4-mini} ~\cite{o3-o4mini-openai-2025}, with other smaller or open-source LLMs: \textit{Deepseek-r1} ~\cite{deepseekai2025deepseekr1incentivizingreasoningcapability}, \textit{gpt-4o-mini} ~\cite{openai2024gpt4ocard}, \textit{Llama3.1:70b} ~\cite{grattafiori2024llama3herdmodels}.

\paragraph{Baseline}
We use prompt-based simulation using Claude Model \textit{claude-3.7-sonnet} \cite{bai2022constitutionalaiharmlessnessai} as our baseline since it has the best performance among all other vanilla LLMs. (see the Appendix for baseline model comparison). Baseline results are produced using only the target turn, a random initial character setting with a brief background and previous experience with no special prompts or parameters, and rely solely on the LLM's ability to generate outputs. 

\paragraph{Evaluation Methods}
For each task, we create a set of simulators and control the parameters to generate task-specific conversations. 

\textit{Topic Diversity} We use a random seed to create 800 entrepreneurs' background data. The generated parameters are then injected into the prompt and fed into each LLM. The evaluation is done by manually eliminating similar topics from the generated results. We also compared the diversity of the topics by entropy: $H(X) = -\sum_{i=1}^np(x_i)\log p(x_i)$, where $p(x_i)$ is the probability topic $x_i$.

\textit{Parameter Adherence} We generate 200 entrepreneurs' background data and randomized conversation parameters. These are fed into each LLM across four different conversation lengths: 5, 10, 15, and 20 turns, resulting in a total of 800 conversations. The evaluation employs a hybrid human-LLM assessment framework in which both human annotators and \textit{Claude-sonnet-3.7} serve as judges.

The evaluation protocol provides judges with only the conversation transcript, requiring them to infer the original parameters based on predefined parameter definitions. For numerical parameters (all on the 1-5 Likert scale), adherence is measured using the mean squared error: $\frac{1}{n}\sum{\text{set value} - \text{inferred value}}$. Categorical parameters are evaluated using multi-class classification accuracy, where correct classifications receive a score of 1 and incorrect classifications receive a score of 0.

To ensure reliability, each conversation is evaluated by both human annotators and the LLM judge. The final parameter adherence score is calculated as the weighted average of human and LLM evaluations, with weights determined by respective agreement levels. Results are reported as MSE for numerical parameters and classification accuracy for categorical parameters, categorized by turns.


\textit{Topic Drift}
We generate 200 20-turn entrepreneur conversations, each with smoothness factor set to A (highest topic adherence) and F (lowest topic adherence), along with 200 baseline conversations without smoothness factor control, resulting in 600 total conversations for topic drift analysis. The smoothness factor parameter controls the degree to which conversations maintain thematic coherence versus allowing natural topic exploration and deviation from the original business concept.

This evaluation measures the semantic distance between conversation segments and the initial topic focus using sentence embedding techniques. We employ BERT-based sentence embeddings to compute cosine similarity scores: $1 - \cos( \text{embedding}(\text{utterance}_i) - \text{embedding}(\text{utterance}_0))$ between the entrepreneur's utterances at each turn and the main business topic established in the conversation opening.

\textit{Character Properties Stability}
We generate 500 20-turn conversations with both the entrepreneur’s formality and technical levels randomized between 0 and 1, then 500 more with the formality parameter omitted and another 500 with the technical parameter omitted.

Character stability is evaluated across the two dimensions: 
\begin{itemize}
\item \textit{Formality Level}: Formality is determined by a composite of vocabulary sophistication, sentence structure, and pronoun usage. 
\item \textit{Technical Language Level}: The technical level is determined by the density of the domain terminology, the complexity of the concepts, and the usage of jargon. 

\noindent The final stability score is calculated by the $1 - 0.5(\text{Formality Error} + \text{Technical Level Error})$
\end{itemize}

\textit{Entity Revisit Rate}
We generate 100 entrepreneurs' background information with Knowledge Gap Level parameters ranging from 1-5, where this parameter measures the knowledge disparity between the user's existing background and their proposed business concept. Each entrepreneur profile is used to generate conversations in four different lengths (5, 10, 15, and 20 turns).

The evaluation is done by first extracting NER and core concepts using BERT. We then track when previously mentioned entities reappear in subsequent turns in the conversation. The concept of a recall rate is calculated as $\frac{1}{T-1}\sum_{t=2}^T |\text{Entities}_t \cap \bigcup_{i=1}^{t-1}\text{Entities}_i|$, where $\text{Entities}_t$ represents the set of entities mentioned at turn $t$, and $T$ is the total duration of the conversation. 

\section{Experiments}
Our experiment results can be summarized as follows.
\begin{table}[t]
  \centering
  \begin{tabular}{lr}
    \hline
    Model         & Embedding diversity \\
    \hline
    claude        & 0.2912 \\
    deepseek‑r1   & 0.4161 \\
    o3            & 0.3360 \\
    o4‑mini       & 0.2830 \\
    gpt‑4.1       & 0.3436 \\
    gpt‑4o‑mini   & 0.2085 \\
    gemini        & 0.3747 \\
    llama3.1:70b        & 0.0576 \\
    baseline      & 0.1075 \\
    \hline
  \end{tabular}
\caption{Embedding diversity (sentence embedding).}
  \label{tab:embed-div}
\end{table}

\begin{table}[t]
  \centering
  \begin{tabular}{lrr}
    \hline
    Model         & Topic diversity & Topic entropy \\
    \hline
    claude        & 111 & 4.469 \\
    deepseek‑r1   & 143 & 5.275 \\
    o3            & 136 & 4.464 \\
    o4‑mini       & 154 & 5.311 \\
    gpt‑4.1       & 140 & 4.578 \\
    gpt‑4o‑mini   &  84 & 3.859 \\
    gemini        & 141 & 5.266 \\
    llama3        &   5 & 0.888 \\
    baseline      &  35 & 2.985 \\
    \hline
  \end{tabular}
  \caption{Topic diversity and topic entropy.}
  \label{tab:topic-div-entropy}
\end{table}

\paragraph{Simulators have bias on topic selection, and may not generate a diverse pool of topics.}
The simulators can be classified into two broad camps according to their approach to exploring the subject matter, as shown in Table \ref{tab:topic-div-entropy}. Advanced models such as \textit{Gemini-2.5-pro} and \textit{DeepSeek-R1} exhibit superior topic diversification capabilities, generating 141 and 143 distinct topics, respectively, with corresponding entropy values of 5.266 and 5.275. These models demonstrate a more uniform attention distribution across thematic domains, closely approximating human-like conversational breadth. In contrast, less capable models like \textit{GPT-4o-mini} produce more constrained topic distributions, while lightweight models such as \textit{Llama3.1:70b} show severe limitations with only 5 distinct topics.

The baseline approach without parameterization yields poor diversity metrics, highlighting the need for structured parameter control. Mid-tier systems occupy an intermediate position, with respectable topic coverage but exhibiting concentration patterns around familiar conceptual clusters. This shows model architectures can explore diverse thematic spaces while maintaining coherent conversational flow.

We also examine sentence diversity by calculating semantic diversity through the cosine similarity of embeddings generated by all-MiniLM-L6-v2. (Table \ref{tab:embed-div}). The embedding diversity rankings partially diverge from topic-level diversity measures, suggesting that models may employ different strategies for achieving variation, and they may use similar words or add additional definitions (e.g., AI-driven business vs. non-AI-driven) to express different topics. 

Beyond quantitative diversity measures, we observe systematic biases in topic selection patterns. For example, when generating food-related business scenarios, models frequently default to vegan or health-conscious options regardless of user specifications. This tendency toward "safe" or socially desirable recommendations indicates inherent training biases that may limit the authenticity of generated conversations. Incorporating structured background parameters significantly reduces these limitations, with all evaluated models showing measurable improvements in topic diversity when provided with detailed entrepreneur profiles.

\begin{figure}[t]
  \centering
  \includegraphics[width=\columnwidth]{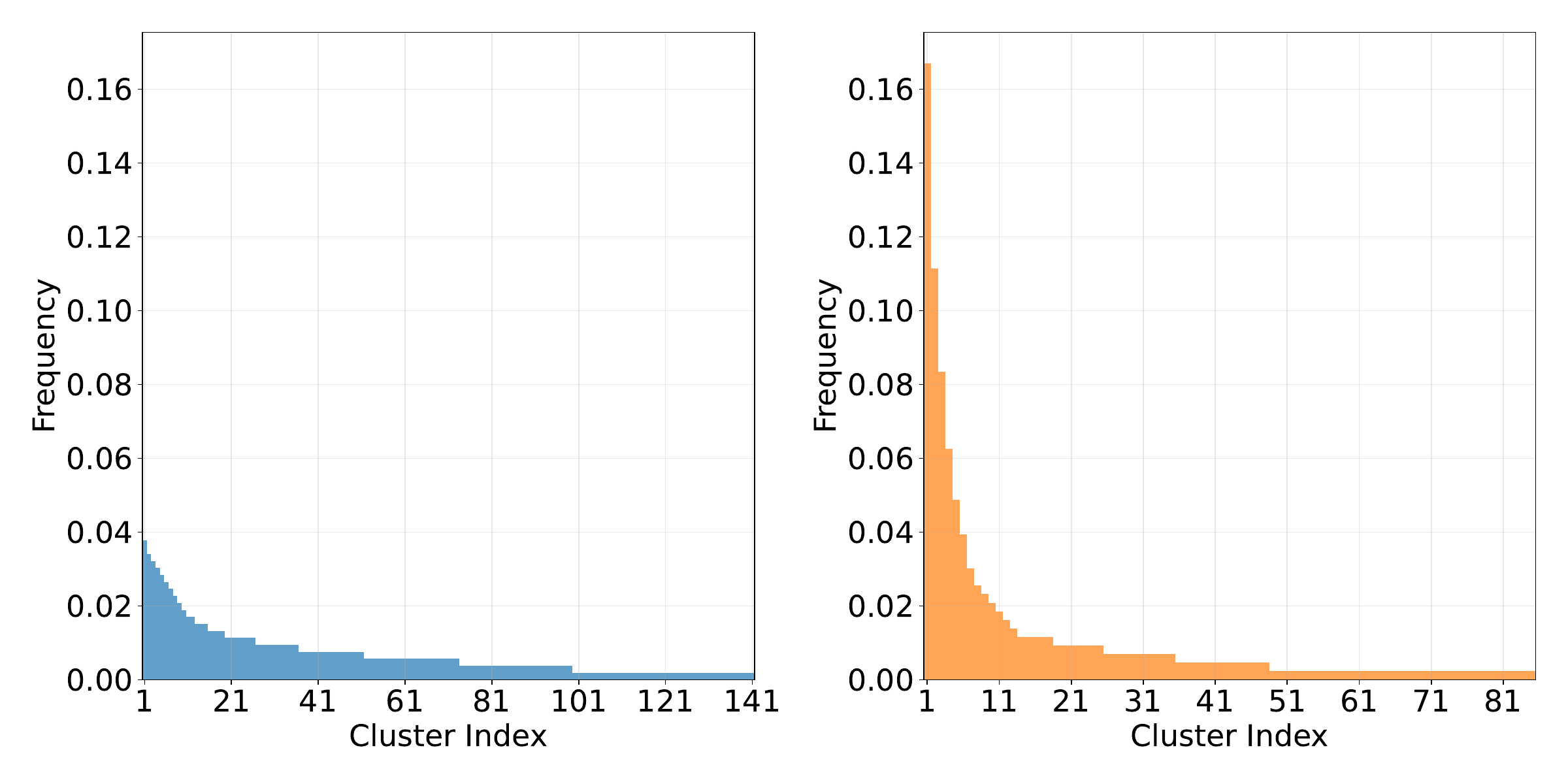}
  \caption{Topic frequency distributions of the 
  \textit{gpt-4o-mini} (orange) and \textit{gemini-2.5-pro} (blue). Clusters are sorted in descending frequency. More advanced model (gemini-2.5-pro) produced a more diverse topic compared to the less advanced model (gpt-4o-mini)}
  \label{fig:distribution-comparison}
\end{figure}

\begin{figure}[t]
  \centering
  \setlength{\tabcolsep}{0pt}
  \begin{tabular}{@{}c@{}}
    \includegraphics[width=0.8\columnwidth]{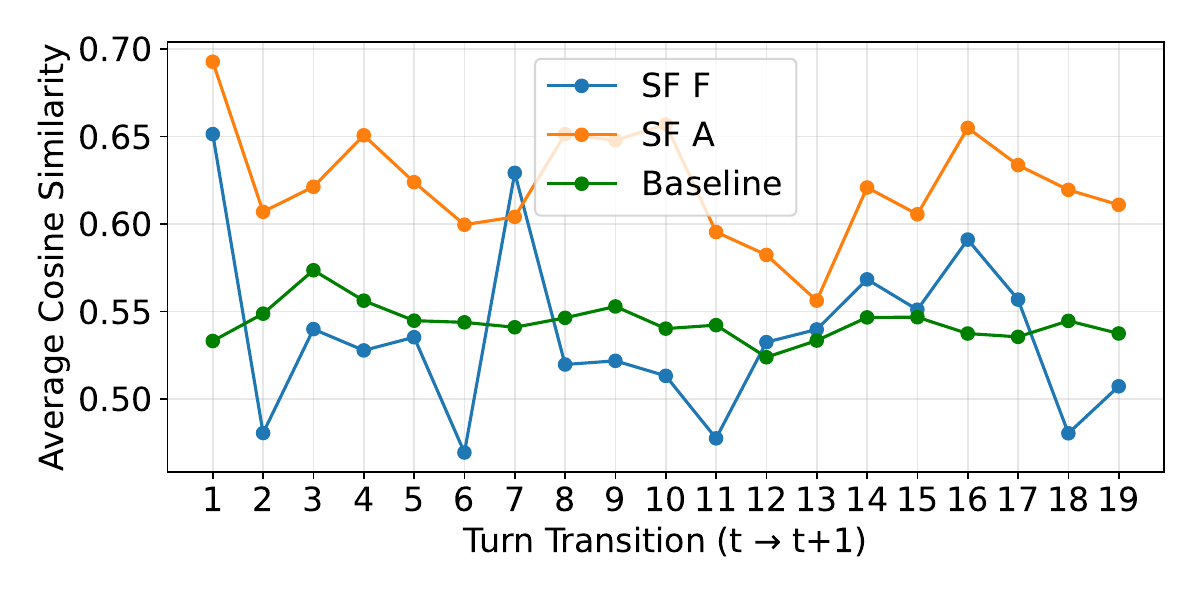} \\
    \includegraphics[width=0.8\columnwidth]{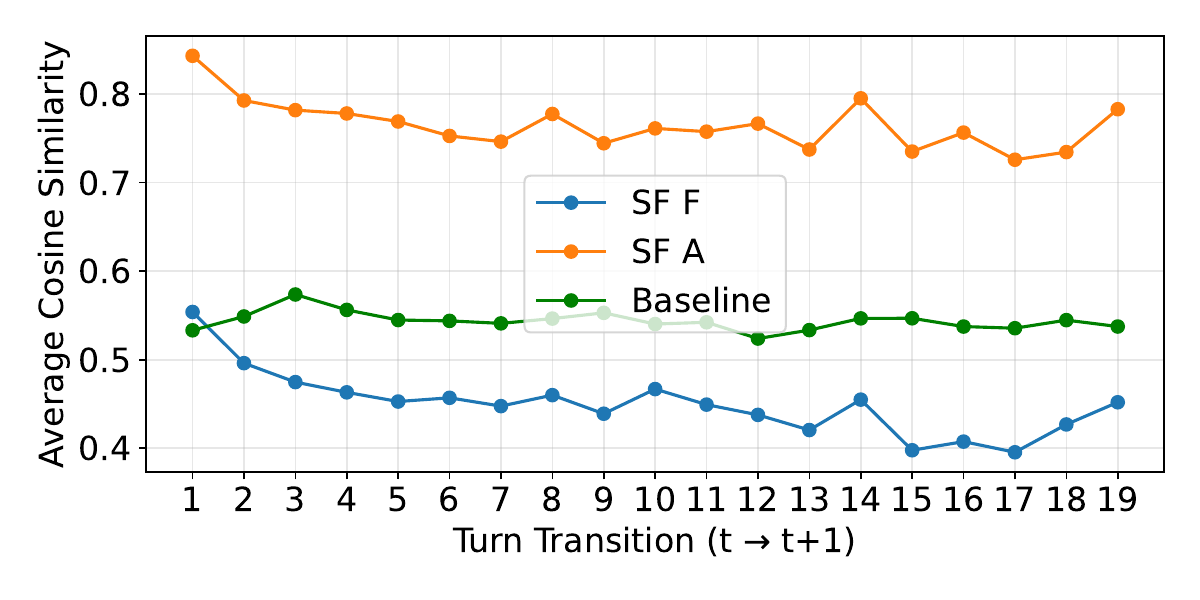} \\
  \end{tabular}
  \caption{The cosine similarity between the entrepreneur's utterance to the main topic in different smoothness factors for two models: (a) \textit{gpt-4o-mini}, (b) \textit{claude-3.7-sonnet}. \textit{claude-3.7-sonnet} is showing a high separation between the highest and lowest smoothness factor, showing better understanding and adherence to parameters.}
  \label{fig:smoothness_factor}
\end{figure}

\paragraph{Adding Smoothness factor improves topic correlation.}
Adding a smoothness factor to simulate the conversation flow not only creates a diversified conversation but also improves the model's adherence to the main topic. (Figure \ref{fig:smoothness_factor}). Both the small and the more advanced models can improve adherence to the main topic after setting a high smoothness factor, and advanced models can successfully create a more significant difference between high and low smoothness factors. Without the smoothness factor, the model can only provide a conversation that has low correlation to the given topic. 

\begin{figure}[t]
  \centering
  \begin{tabular}{@{}cc@{}}
    \includegraphics[width=0.48\columnwidth]{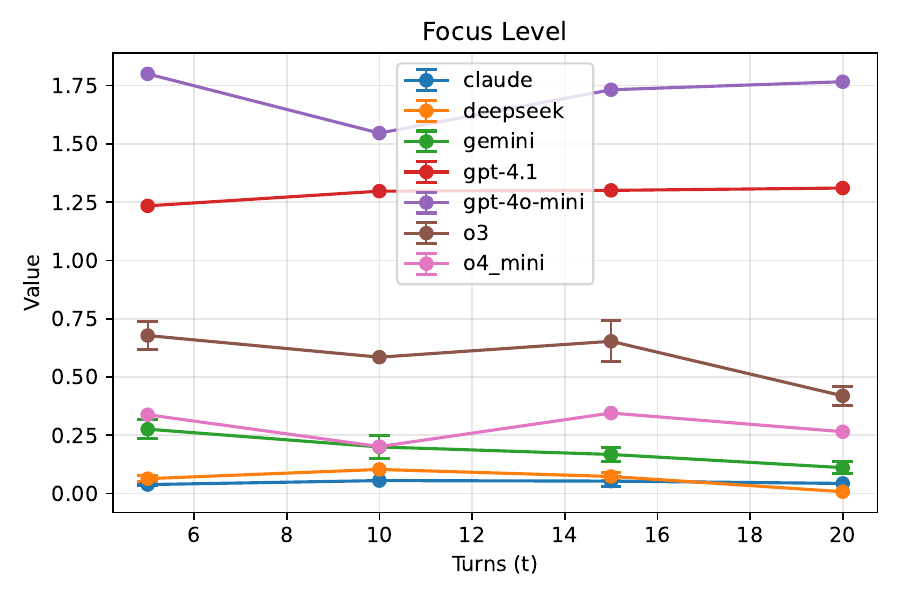} &
    \includegraphics[width=0.48\columnwidth]{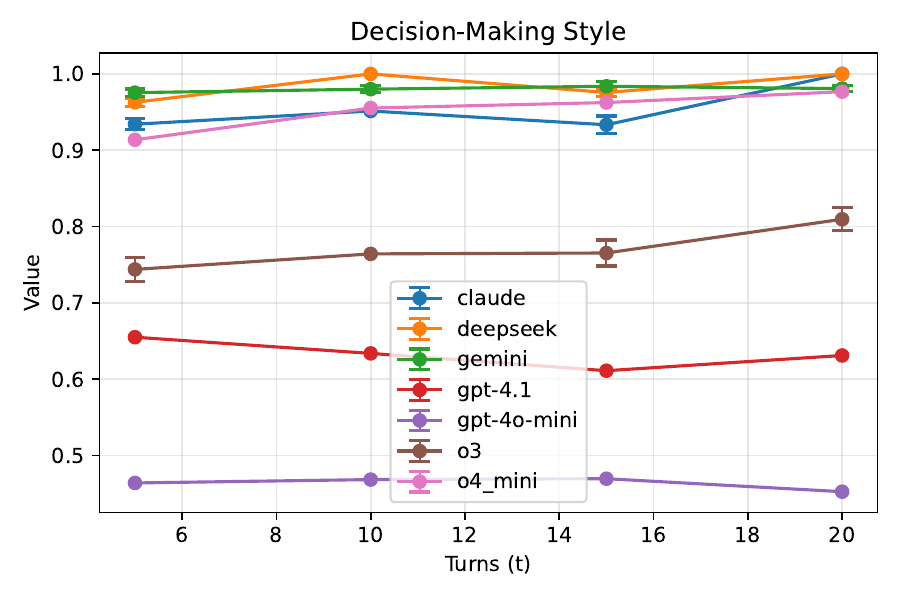} \\
    (a) Focus Level & (b) Decision‑Making Style \\[8pt]
    \includegraphics[width=0.48\columnwidth]{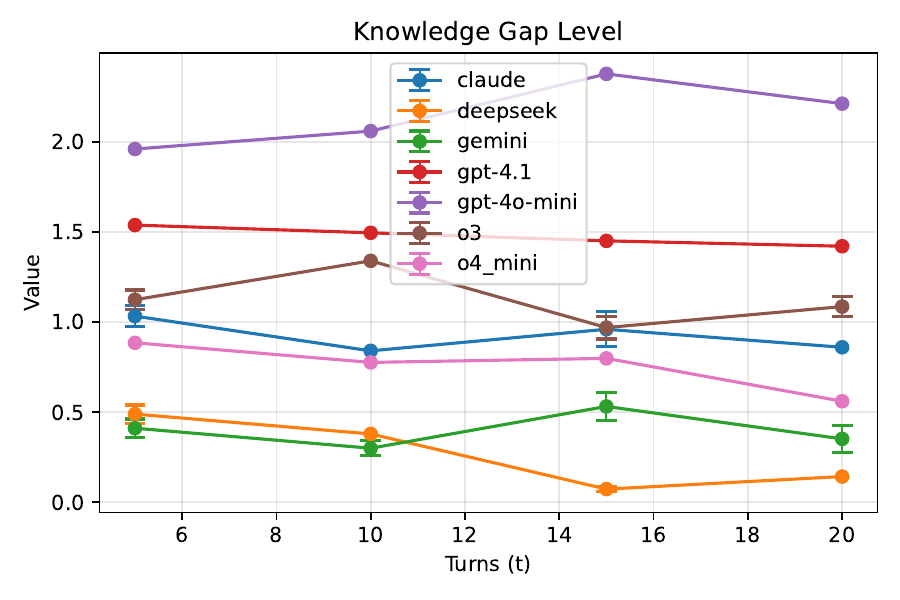} &
    \includegraphics[width=0.48\columnwidth]{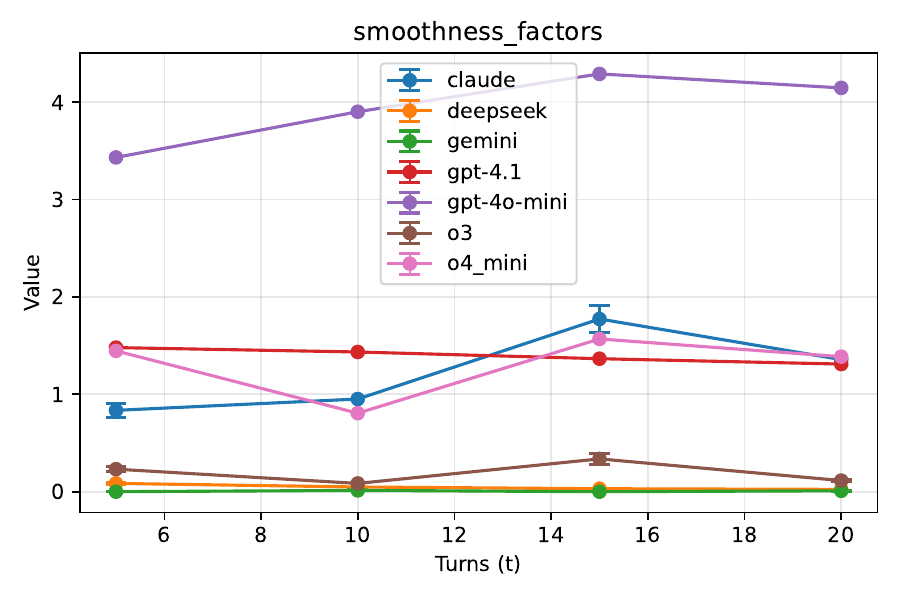} \\
    (c) Knowledge Gap Level & (d) Smoothness Factors
  \end{tabular}
  \caption{Model metric curves vs.\ conversation turns.}
  \label{fig:metrics-2x2}
\end{figure}

\paragraph{Parameter adherence varies across models with improving accuracy over extended conversations.}
Analysis of parameter adherence across conversation turns reveals substantial differences in model capabilities, with most parameters showing improved accuracy as conversations progress. As shown in Figures \ref{fig:metrics-2x2}(a-c), advanced models such as Claude and Gemini demonstrate superior parameter implementation, with MSE errors for the focus level, the knowledge gap level, and the experience level decreasing from initial values to more accurate parameter representation over 20 conversations. This improvement pattern suggests that models require several turns to fully establish and maintain specified parameter values. The evaluation of the decision-making style (Figure \ref{fig:metrics-2x2} (b)) shows binary classification accuracy, where advanced models achieve 0.8-1.0 accuracy rates while lighter models like \textit{gpt-4o-mini} struggle to maintain consistent classification performance, often hovering around 0.4-0.6 accuracy. The smoothness factor analysis (Figure \ref{fig:metrics-2x2}(d)) demonstrates that parameter control effectiveness varies significantly by model architecture, with Claude maintaining clear parameter differentiation while smaller models show less distinct parameter implementation regardless of specified values.

\paragraph{Knowledge gap parameters influence concept revisit patterns in advanced models.}
The relationship between Knowledge Gap Level and concept revisit behavior reveals substantial differences in advanced models' adaptation capabilities, as shown in Figure \ref{fig:revisit_vertical_tabular}. \textit{Gemini-2.5-pro} exhibits a clear inverse correlation between knowledge gap and revisit rate, with highly knowledgeable users (Level 1) showing revisit rates of approximately 0.5-0.6, while novice users (Level 5) demonstrate lower revisit rates around 0.1-0.2 across all conversation lengths. This pattern aligns with pedagogical theory, where experts benefit from reinforcement of complex concepts, while beginners require more linear information introduction. Conversely, Claude shows a lower differentiation between knowledge gap levels, but a higher differentiation over turns. This shows that some models cannot correctly simulate a conversation with low revisit rates.

\begin{figure}[t]
  \centering
  \setlength{\tabcolsep}{0pt}
    \includegraphics[width=0.8\columnwidth]{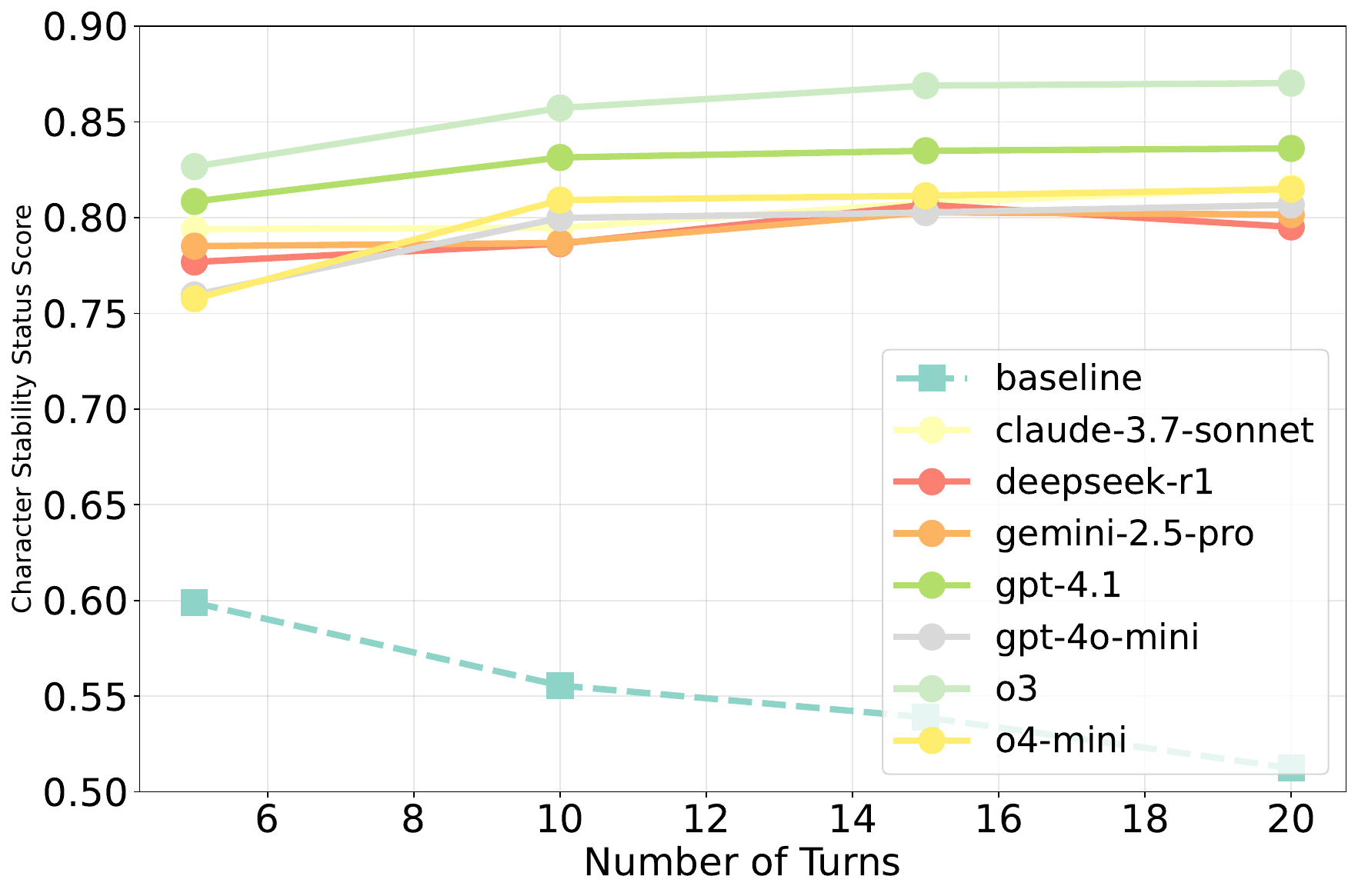} \\
  \caption{Character Parameter stability over turns, the baseline has a decreasing stability over turns, while all other models with character properties show an increase in stability score.}
  \label{fig:coref-vertical}
\end{figure}

With a high knowledge gap level, all models show a higher revisit rate compared to the baseline. (Figure \ref{fig:coref-vertical}). Advanced models, including \textit{o3}, \textit{gpt-4.1}, and \textit{Claude-3.7-sonnet}, maintain high character consistency scores that improve over extended conversations, while mid-tier models show respectable but more variable performance. The baseline approach demonstrates significantly lower consistency. This suggests that sophisticated parameter implementation requires substantial model capacity to fully understand and adhere to the parameters, but all models can obtain a significant level of performance increase.

\begin{figure}
    \centering
    \includegraphics[width=\linewidth]{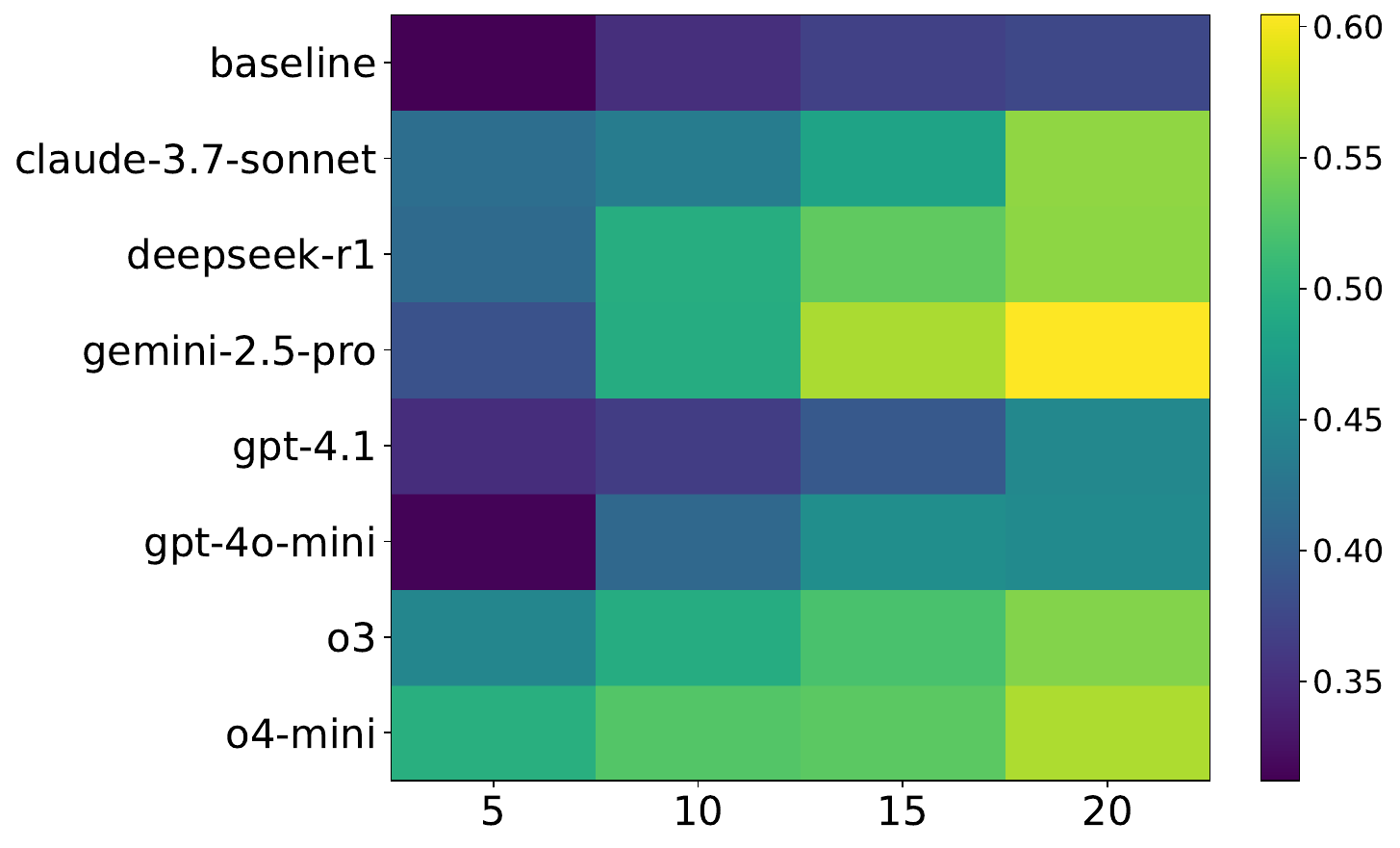}
    \caption{Concept-revisit rate by turns for each model with knowledge gap level of user set to 1 (most knowledgeable). All models exhibit a higher revisit rate with turn progression.}
    \label{fig:revisit_heatmap}
\end{figure}

\begin{figure}[t]
  \centering
  \setlength{\tabcolsep}{0pt}
  \begin{tabular}{@{}c@{}}
    \includegraphics[width=0.8\columnwidth]{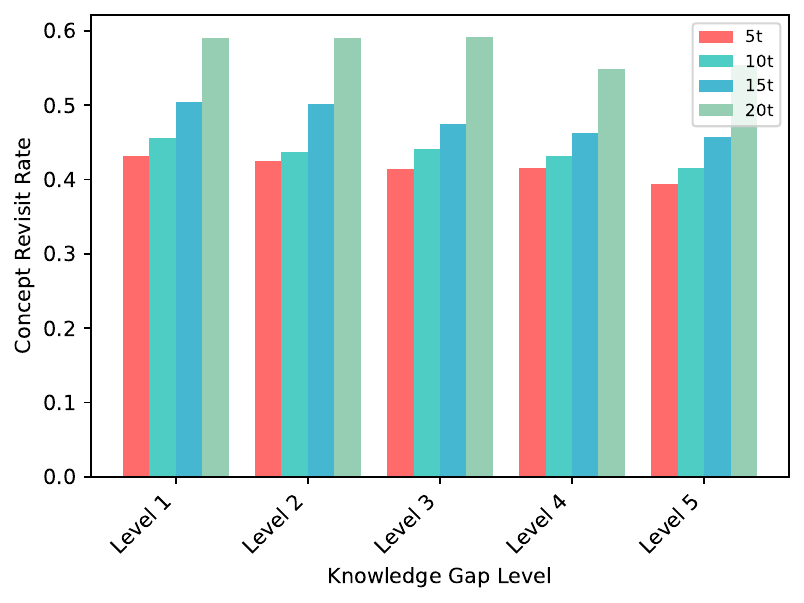} \\
    \includegraphics[width=0.8\columnwidth]{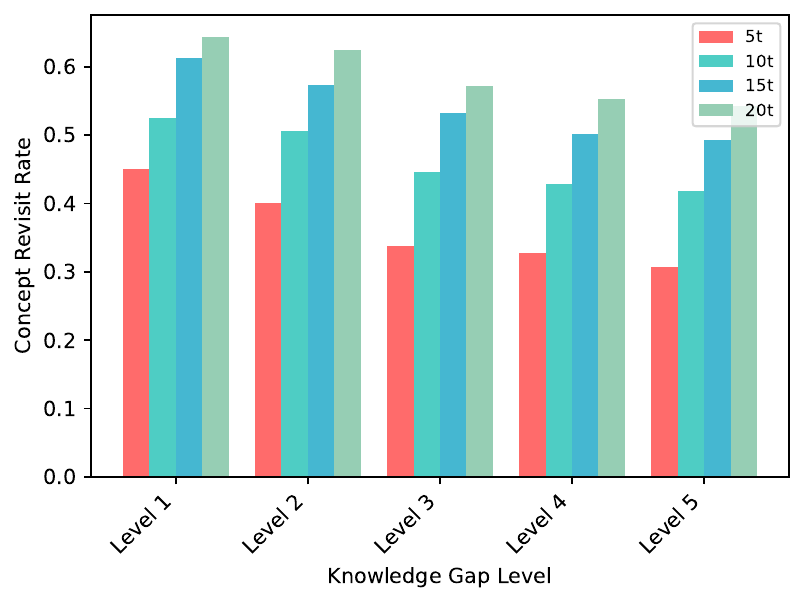} \\
  \end{tabular}
  \caption{Concept‑revisit rate by knowledge‑gap level for two models: (a) Claude, (b) Gemini. Knowledge Gap Level 1 is the smallest knowledge gap, and Knowledge Gap Level 5 is the highest. \textit{Gemini-2.5-pro} shows a more significant difference when modifying Knowledge Gap Level.}
  \label{fig:revisit_vertical_tabular}
\end{figure}

\paragraph{Character parameters are stable across all models.}
The character parameter study shows that all models can reach high parameter stability over turns, although more advanced models have better performance (Figure \ref{fig:coref-vertical}). All models exhibit improved stability trajectories over conversation length, with consistency scores rising from initial values. This could be because the model does not have enough context initially, but the performance stabilizes after 15 turns. 

We also performed an ablation analysis presented in Table \ref{tab:perf-turns}, where we test the error of the model when only the formality parameters of the model or technical parameters are given. The result shows that the combined parameter implementation yields benefits exceeding the sum of individual components in both models. This suggests that adding more specified parameters to the model may further increase the model's capability of simulating complex conversations. 

\begin{table}[t]
  \centering
  \small
  \begin{tabular}{lrrrr}
    \hline
    Model & Turns & Formality & Technical & Full \\
    \hline
    claude-3.7-sonnet & 5  & 0.280 & 0.252 & 0.206 \\
    claude-3.7-sonnet & 10 & 0.305 & 0.265 & 0.205 \\
    claude-3.7-sonnet & 15 & 0.298 & 0.258 & 0.192 \\
    claude-3.7-sonnet & 20 & 0.292 & 0.252 & 0.184 \\
    o3 & 5 & 0.255 & 0.212 & 0.173\\
    o3 & 10 & 0.222 & 0.175 & 0.143\\
    o3 & 15 & 0.215 & 0.162 & 0.131\\
    o3 & 20 & 0.212 & 0.155 & 0.130\\
    \hline
  \end{tabular}
  \caption{Average performance errors for Formality Only, Technical Only, and Full Parameters across varying conversation turns.}
  \label{tab:perf-turns}
\end{table}

\begin{figure}[ht]
  \centering
  \setlength{\tabcolsep}{0pt}
    \includegraphics[width=0.8\columnwidth]{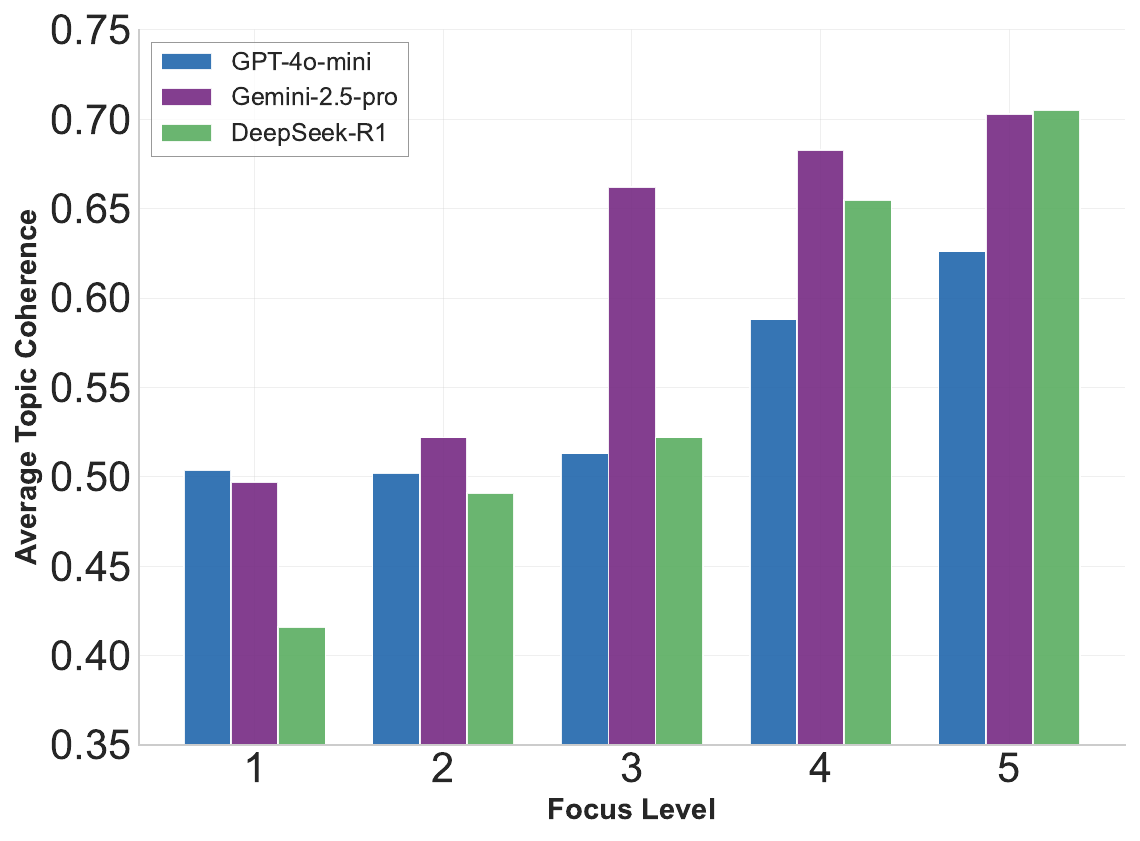} \\
  \caption{Average topic coherence between turns, models have parameter sensitivity issues on intermediate values, but all models can differentiate the lowest and highest value.}
  \label{fig:topic-coherence}
\end{figure}
\paragraph{While simulators can generate good responses, they may fail to create bad ones}

While models demonstrate clear differentiation between extreme parameter values in focus levels (Level 1 vs Level 5), they exhibit poor sensitivity to intermediate parameter settings. In Figure \ref{fig:topic-coherence}, all three models show relatively flat performance curves across the middle range (Levels 2-4), with topic coherence scores clustering around 0.45-0.55 regardless of the specified focus level. This suggests that models can successfully implement "very focused" versus "very unfocused" conversation styles but struggle to generate nuanced variations in between.

Similar behavior is observed in Figures \ref{fig:smoothness_factor} and \ref{fig:coref-vertical}. In Figure \ref{fig:smoothness_factor}, both models show only marginally lower cosine similarity scores compared to the baseline, failing to achieve the expected degradation specified by smoothness factor F (\textit{Highly disjointed with random topic jumping}). In Figure \ref{fig:coref-vertical}, \textit{claude-3.7-sonnet} demonstrates minimal differentiation between knowledge gap levels 1 and 5, while \textit{gemini-2.5-pro} exhibits comparable limitations, conflating performance across levels 3-5 despite maintaining clear separation between the extreme values (levels 1 and 5).

The insensitivity of the parameter may be due to a lack of fine-tuning. With only the definitions for each level provided to the LLM, models can only rely on pre-trained representations to map abstract parameter descriptions to concrete output behaviors. Given sufficient examples of intermediate quality levels between "highly focused" and "completely unfocused" conversations, the model could possibly provide a more distinguishable result. Further, post-training alignment procedures through RLHF further reinforce the model's tendency to produce helpful, coherent responses, creating systematic resistance to generating lower-quality content regardless of parameter specifications, which lowers the model's ability to generate poor-quality conversations.

\section{Conclusion and Discussion}
We create a comprehensive parameterization framework for controlling LLM-based conversation generation, demonstrating both the potential and limitations of current approaches to fine-grained dialogue control. Our experiments with the simulator show that advanced models can effectively differentiate between parameter values and maintain improving consistency over long conversations. 

However, several issues are unaddressed in this exploratory study. We only provide the necessary parameters for conversation generation, not an exhaustive set of parameters that covers all aspects. More parameters could be added to the prompt since we have already proven that interconnected parameters can improve conversation quality. 

A fine-tuned LLM with human-labeled conversation parameters as a dataset may increase the simulator's sensitivity to intermediate values. We are using the default temperature settings. More analysis could be made on different parameter settings and fine-tuned open-source LLMs.

Parameterized settings cannot increase the model's factual accuracy. Adding a factual accuracy parameter can prompt the LLM to provide incorrect information, but they are also not sensitive enough to intermediate parameters and does not decrease the hallucination rate compared to the vanilla model. A RAG-based approach is still needed to decrease the simulator's hallucination.

\newpage

\bibliography{aaai2026}

\begin{thebibliography}{67}
\providecommand{\natexlab}[1]{#1}

\bibitem[{Agostinelli et~al.(2023)Agostinelli, Denk, Borsos, Engel, Verzetti et~al.}]{agostinelli2023musiclmgeneratingmusictext}
Agostinelli, A.; Denk, T.~I.; Borsos, Z.; Engel, J.; Verzetti, M.; et~al. 2023.
\newblock MusicLM: Generating Music From Text.
\newblock arXiv:2301.11325.

\bibitem[{Aher, Arriaga, and Kalai(2023)}]{aher2023using}
Aher, G.~V.; Arriaga, R.~I.; and Kalai, A.~T. 2023.
\newblock Using large language models to simulate multiple humans and replicate human subject studies.
\newblock In \emph{International conference on machine learning}, 337--371. PMLR.

\bibitem[{Anderson et~al.(2001)Anderson, Krathwohl, Airasian, Cruikshank, Mayer, Pintrich, Raths, and Wittrock}]{anderson2001taxonomy}
Anderson, L.~W.; Krathwohl, D.~R.; Airasian, P.~W.; Cruikshank, K.~A.; Mayer, R.~E.; Pintrich, P.~R.; Raths, J.; and Wittrock, M.~C. 2001.
\newblock A Taxonomy for Learning, Teaching, and Assessing: A Revision of Bloom's Taxonomy of Educational Objectives.
\newblock \emph{Educational Horizons}, 83(3): 154--159.

\bibitem[{Austin(1975)}]{austin1975things}
Austin, J.~L. 1975.
\newblock \emph{How to Do Things with Words}.
\newblock Oxford University Press.

\bibitem[{Bai et~al.(2022)Bai, Kadavath, Kundu, Askell, Kernion, Jones, Chen, Goldie et~al.}]{bai2022constitutionalaiharmlessnessai}
Bai, Y.; Kadavath, S.; Kundu, S.; Askell, A.; Kernion, J.; Jones, A.; Chen, A.; Goldie, A.; et~al. 2022.
\newblock Constitutional AI: Harmlessness from AI Feedback.
\newblock arXiv:2212.08073.

\bibitem[{Bamman, O'Connor, and Smith(2013)}]{bamman2013learning}
Bamman, D.; O'Connor, B.; and Smith, N.~A. 2013.
\newblock Learning Latent Personas of Film Characters.
\newblock In \emph{Proceedings of the 51st Annual Meeting of the Association for Computational Linguistics}, 352--361.

\bibitem[{Bamman, Underwood, and Smith(2014)}]{bamman2014bayesian}
Bamman, D.; Underwood, T.; and Smith, N.~A. 2014.
\newblock A Bayesian Mixed Effects Model of Literary Character.
\newblock In \emph{Proceedings of the 52nd Annual Meeting of the Association for Computational Linguistics}, 370--379.

\bibitem[{Baskar et~al.(2025)Baskar, Verelakar, Parthasarathy, and Gaur}]{baskar2025guessingaskingapproachresolving}
Baskar, S.; Verelakar, T.~T.; Parthasarathy, S.; and Gaur, M. 2025.
\newblock From Guessing to Asking: An Approach to Resolving the Persona Knowledge Gap in LLMs during Multi-Turn Conversations.
\newblock arXiv:2503.12556.

\bibitem[{Bender et~al.(2021)Bender, Gebru, McMillan-Major, and Shmitchell}]{bender2021dangers}
Bender, E.~M.; Gebru, T.; McMillan-Major, A.; and Shmitchell, S. 2021.
\newblock On the Dangers of Stochastic Parrots: Can Language Models Be Too Big?
\newblock In \emph{Proceedings of the 2021 ACM Conference on Fairness, Accountability, and Transparency}, 610--623.

\bibitem[{Bloom(1956)}]{bloom1956taxonomy}
Bloom, B.~S. 1956.
\newblock \emph{Taxonomy of Educational Objectives: The Classification of Educational Goals}.
\newblock Longmans, Green.

\bibitem[{Chen et~al.(2021)Chen, Tworek, Jun, Yuan, de~Oliveira~Pinto, Kaplan, Edwards, Burda, Joseph, Brockman, Ray, Puri, Krueger, Petrov et~al.}]{chen2021evaluatinglargelanguagemodels}
Chen, M.; Tworek, J.; Jun, H.; Yuan, Q.; de~Oliveira~Pinto, H.~P.; Kaplan, J.; Edwards, H.; Burda, Y.; Joseph, N.; Brockman, G.; Ray, A.; Puri, R.; Krueger, G.; Petrov, M.; et~al. 2021.
\newblock Evaluating Large Language Models Trained on Code.
\newblock arXiv:2107.03374.

\bibitem[{Comanici et~al.(2025)Comanici, Bieber, Schaekermann, Pasupat, Sachdeva, Dhillon, Blistein, Ram et~al.}]{comanici2025gemini25pushingfrontier}
Comanici, G.; Bieber, E.; Schaekermann, M.; Pasupat, I.; Sachdeva, N.; Dhillon, I.; Blistein, M.; Ram, O.; et~al. 2025.
\newblock Gemini 2.5: Pushing the Frontier with Advanced Reasoning, Multimodality, Long Context, and Next Generation Agentic Capabilities.
\newblock arXiv:2507.06261.

\bibitem[{Copet et~al.(2023)Copet, Kreuk, Gat, Remez, Kant, Synnaeve, Adi, and D{\'e}fossez}]{copet2023simple}
Copet, J.; Kreuk, F.; Gat, I.; Remez, T.; Kant, D.; Synnaeve, G.; Adi, Y.; and D{\'e}fossez, A. 2023.
\newblock Simple and controllable music generation.
\newblock \emph{Advances in Neural Information Processing Systems}, 36: 47704--47720.

\bibitem[{Dathathri et~al.(2019)Dathathri, Madotto, Lan, Hung, Frank, Molino, Yosinski, and Liu}]{dathathri2019plug}
Dathathri, S.; Madotto, A.; Lan, J.; Hung, J.; Frank, E.; Molino, P.; Yosinski, J.; and Liu, R. 2019.
\newblock Plug and Play Language Models: A Simple Approach to Controlled Text Generation.
\newblock In \emph{International Conference on Learning Representations}.

\bibitem[{DeepSeek-AI et~al.(2025)DeepSeek-AI, Guo, Yang, Zhang, Song, Zhang, Xu, Zhu, Ma, Wang et~al.}]{deepseekai2025deepseekr1incentivizingreasoningcapability}
DeepSeek-AI; Guo, D.; Yang, D.; Zhang, H.; Song, J.; Zhang, R.; Xu, R.; Zhu, Q.; Ma, S.; Wang, P.; et~al. 2025.
\newblock DeepSeek-R1: Incentivizing Reasoning Capability in LLMs via Reinforcement Learning.
\newblock arXiv:2501.12948.

\bibitem[{Deriu et~al.(2020)Deriu, Rodrigo, Otegi, Echegoyen, Rosset, Agirre, and Cieliebak}]{deriu2020survey}
Deriu, J.; Rodrigo, A.; Otegi, A.; Echegoyen, G.; Rosset, S.; Agirre, E.; and Cieliebak, M. 2020.
\newblock Survey on evaluation methods for dialogue systems.
\newblock \emph{Artificial Intelligence Review}, 54(1): 755--810.

\bibitem[{Drori et~al.(2022)Drori, Kharkar, Sickinger, Kochman, Ng, Hadash, Ge, Tenenbaum, Liu, and Strobelt}]{drori2022neural}
Drori, I.; Kharkar, A.; Sickinger, W.~R.; Kochman, E.; Ng, S.~P.; Hadash, K.; Ge, Y.; Tenenbaum, J.~B.; Liu, C.; and Strobelt, H. 2022.
\newblock A Neural Network Solves, Explains, and Generates University Math Problems by Program Synthesis and Few-Shot Learning at Human Level.
\newblock In \emph{Proceedings of the National Academy of Sciences}, volume 119, e2123433119.

\bibitem[{Dziri et~al.(2020)Dziri, Kamalloo, Mathewson, and Zaiane}]{dziri2020evaluatingcoherencedialoguesystems}
Dziri, N.; Kamalloo, E.; Mathewson, K.~W.; and Zaiane, O. 2020.
\newblock Evaluating Coherence in Dialogue Systems using Entailment.
\newblock arXiv:1904.03371.

\bibitem[{Dziri et~al.(2022)Dziri, Kamalloo, Milton, Zaiane, Yu, Ponti, and Reddy}]{dziri2022origin}
Dziri, N.; Kamalloo, E.; Milton, S.; Zaiane, O.; Yu, M.; Ponti, E.; and Reddy, S. 2022.
\newblock On the Origin of Hallucinations in Conversational Models: Is it the Datasets or the Models?
\newblock In \emph{Proceedings of the 2022 Conference of the North American Chapter of the Association for Computational Linguistics: Human Language Technologies}, 5271--5285.

\bibitem[{Gao, Galley, and Li(2018)}]{gao2018neural}
Gao, J.; Galley, M.; and Li, L. 2018.
\newblock Neural approaches to conversational AI.
\newblock In \emph{The 41st international ACM SIGIR conference on research \& development in information retrieval}, 1371--1374.

\bibitem[{Grattafiori et~al.(2024)Grattafiori, Dubey, Jauhri, Pandey, Kadian, Al-Dahle, Letman, Mathur et~al.}]{grattafiori2024llama3herdmodels}
Grattafiori, A.; Dubey, A.; Jauhri, A.; Pandey, A.; Kadian, A.; Al-Dahle, A.; Letman, A.; Mathur, A.; et~al. 2024.
\newblock The Llama 3 Herd of Models.
\newblock arXiv:2407.21783.

\bibitem[{Hu et~al.(2017)Hu, Yang, Liang, Salakhutdinov, and Xing}]{hu2017toward}
Hu, Z.; Yang, Z.; Liang, X.; Salakhutdinov, R.; and Xing, E.~P. 2017.
\newblock Toward Controlled Generation of Text.
\newblock In \emph{Proceedings of the 34th International Conference on Machine Learning}, volume~70, 1587--1596.

\bibitem[{Huang, Zhu, and Gao(2020)}]{huang2020challenges}
Huang, M.; Zhu, X.; and Gao, J. 2020.
\newblock Challenges in Building Intelligent Open-domain Dialog Systems.
\newblock In \emph{ACM Transactions on Information Systems}, volume~38, 1--32.

\bibitem[{Jumper et~al.(2021)Jumper, Evans, Pritzel, Green, Figurnov, Ronneberger, Tunyasuvunakool, Bates, Žídek, Potapenko et~al.}]{jumper2021highly}
Jumper, J.; Evans, R.; Pritzel, A.; Green, T.; Figurnov, M.; Ronneberger, O.; Tunyasuvunakool, K.; Bates, R.; Žídek, A.; Potapenko, A.; et~al. 2021.
\newblock Highly Accurate Protein Structure Prediction with {AlphaFold}.
\newblock \emph{Nature}, 596(7873): 583--589.

\bibitem[{Jurafsky and Martin(2000)}]{jurafsky2000speech}
Jurafsky, D.; and Martin, J.~H. 2000.
\newblock \emph{Speech and Language Processing: An Introduction to Natural Language Processing, Computational Linguistics, and Speech Recognition}.
\newblock Prentice Hall.

\bibitem[{Keskar et~al.(2019)Keskar, McCann, Varshney, Xiong, and Socher}]{keskar2019ctrlconditionaltransformerlanguage}
Keskar, N.~S.; McCann, B.; Varshney, L.~R.; Xiong, C.; and Socher, R. 2019.
\newblock CTRL: A Conditional Transformer Language Model for Controllable Generation.
\newblock arXiv:1909.05858.

\bibitem[{Khalifa, Elsahar, and Dymetman(2021)}]{khalifa2021distributional}
Khalifa, M.; Elsahar, H.; and Dymetman, M. 2021.
\newblock A Distributional Approach to Controlled Text Generation.
\newblock In \emph{International Conference on Learning Representations}.

\bibitem[{Kim, Soyata, and Behnagh(2020)}]{kim2020designing}
Kim, Y.; Soyata, T.; and Behnagh, R.~F. 2020.
\newblock Designing adaptive conversational agent tutors to enhance computer science learning through dialogue-based scaffolding.
\newblock \emph{Computers in Human Behavior}, 113: 106496.

\bibitem[{Larochelle et~al.(2009)Larochelle, Bengio, Louradour, and Lamblin}]{larochelle2009exploring}
Larochelle, H.; Bengio, Y.; Louradour, J.; and Lamblin, P. 2009.
\newblock Exploring strategies for training deep neural networks.
\newblock \emph{Journal of machine learning research}, 10(1).

\bibitem[{Larsson and Traum(2000)}]{larsson2000information}
Larsson, S.; and Traum, D.~R. 2000.
\newblock Information State and Dialogue Management in the {TRINDI} Dialogue Move Engine Toolkit.
\newblock \emph{Natural Language Engineering}, 6(3-4): 323--340.

\bibitem[{Lester, Al-Rfou, and Constant(2021)}]{lester2021powerscaleparameterefficientprompt}
Lester, B.; Al-Rfou, R.; and Constant, N. 2021.
\newblock The Power of Scale for Parameter-Efficient Prompt Tuning.
\newblock arXiv:2104.08691.

\bibitem[{Li et~al.(2016{\natexlab{a}})Li, Galley, Brockett, Gao, and Dolan}]{li2016diversitypromotingobjectivefunctionneural}
Li, J.; Galley, M.; Brockett, C.; Gao, J.; and Dolan, B. 2016{\natexlab{a}}.
\newblock A Diversity-Promoting Objective Function for Neural Conversation Models.
\newblock arXiv:1510.03055.

\bibitem[{Li et~al.(2016{\natexlab{b}})Li, Galley, Brockett, Spithourakis, Gao, and Dolan}]{li2016persona}
Li, J.; Galley, M.; Brockett, C.; Spithourakis, G.; Gao, J.; and Dolan, B. 2016{\natexlab{b}}.
\newblock A Persona-Based Neural Conversation Model.
\newblock In \emph{Proceedings of the 54th Annual Meeting of the Association for Computational Linguistics}, 994--1003.

\bibitem[{Li and Liang(2021)}]{li2021prefixtuningoptimizingcontinuousprompts}
Li, X.~L.; and Liang, P. 2021.
\newblock Prefix-Tuning: Optimizing Continuous Prompts for Generation.
\newblock arXiv:2101.00190.

\bibitem[{Li et~al.(2023)Li, Li, Zhang, Dan, and Zhang}]{tan2023chatdoctor}
Li, Y.; Li, Z.; Zhang, K.; Dan, R.; and Zhang, Y. 2023.
\newblock ChatDoctor: A Medical Chat Model Fine-tuned on a Large Language Model Meta-AI (LLaMA) using Medical Domain Knowledge.
\newblock \emph{arXiv preprint arXiv:2303.14070}.

\bibitem[{Lowe et~al.(2018)Lowe, Noseworthy, Serban, Angelard-Gontier, Bengio, and Pineau}]{lowe2018automaticturingtestlearning}
Lowe, R.; Noseworthy, M.; Serban, I.~V.; Angelard-Gontier, N.; Bengio, Y.; and Pineau, J. 2018.
\newblock Towards an Automatic Turing Test: Learning to Evaluate Dialogue Responses.
\newblock arXiv:1708.07149.

\bibitem[{Mairesse et~al.(2007)Mairesse, Walker, Mehl, and Moore}]{mairesse2007using}
Mairesse, F.; Walker, M.~A.; Mehl, M.~R.; and Moore, R.~K. 2007.
\newblock Using Linguistic Cues for the Automatic Recognition of Personality in Conversation and Text.
\newblock In \emph{Journal of Artificial Intelligence Research}, volume~30, 457--500.

\bibitem[{Mani(2012)}]{mani2012computational}
Mani, I. 2012.
\newblock \emph{Computational Modeling of Narrative}.
\newblock Morgan \& Claypool Publishers.

\bibitem[{Mehri and Eskenazi(2020)}]{mehri2020usrunsupervisedreferencefree}
Mehri, S.; and Eskenazi, M. 2020.
\newblock USR: An Unsupervised and Reference Free Evaluation Metric for Dialog Generation.
\newblock arXiv:2005.00456.

\bibitem[{Min et~al.(2022)Min, Lyu, Holtzman, Artetxe, Lewis, Hajishirzi, and Zettlemoyer}]{min2022rethinking}
Min, S.; Lyu, X.; Holtzman, A.; Artetxe, M.; Lewis, M.; Hajishirzi, H.; and Zettlemoyer, L. 2022.
\newblock Rethinking the Role of Demonstrations: What Makes In-Context Learning Work?
\newblock In \emph{Proceedings of the 2022 Conference on Empirical Methods in Natural Language Processing}, 7102--7113.

\bibitem[{Mishra et~al.(2022)Mishra, Khashabi, Baral, and Hajishirzi}]{mishra2022reframing}
Mishra, S.; Khashabi, D.; Baral, C.; and Hajishirzi, H. 2022.
\newblock Reframing Instructional Prompts to {GPT}k's Language.
\newblock \emph{Findings of the Association for Computational Linguistics: ACL 2022}, 589--612.

\bibitem[{Nichol et~al.(2022)Nichol, Dhariwal, Ramesh, Shyam, Mishkin, McGrew, Sutskever, and Chen}]{nichol2022glidephotorealisticimagegeneration}
Nichol, A.; Dhariwal, P.; Ramesh, A.; Shyam, P.; Mishkin, P.; McGrew, B.; Sutskever, I.; and Chen, M. 2022.
\newblock GLIDE: Towards Photorealistic Image Generation and Editing with Text-Guided Diffusion Models.
\newblock arXiv:2112.10741.

\bibitem[{Nye, Graesser, and Hu(2014)}]{nye2014autotutor}
Nye, B.~D.; Graesser, A.~C.; and Hu, X. 2014.
\newblock {AutoTutor} and Family: A Review of 17 Years of Natural Language Tutoring.
\newblock \emph{International Journal of Artificial Intelligence in Education}, 24(4): 427--469.

\bibitem[{OpenAI et~al.(2024{\natexlab{a}})OpenAI, :, Hurst, Lerer, Goucher, Perelman, Ramesh, Clark, Ostrow, Welihinda, Hayes, Radford, Mądry, Baker-Whitcomb, Beutel et~al.}]{openai2024gpt4ocard}
OpenAI; :; Hurst, A.; Lerer, A.; Goucher, A.~P.; Perelman, A.; Ramesh, A.; Clark, A.; Ostrow, A.; Welihinda, A.; Hayes, A.; Radford, A.; Mądry, A.; Baker-Whitcomb, A.; Beutel, A.; et~al. 2024{\natexlab{a}}.
\newblock GPT-4o System Card.
\newblock arXiv:2410.21276.

\bibitem[{OpenAI(2025)}]{o3-o4mini-openai-2025}
OpenAI. 2025.
\newblock {Introducing OpenAI o3 and o4‑mini}.
\newblock \url{https://openai.com/index/introducing-o3-and-o4-mini/}.
\newblock Accessed: 2025-07-29.

\bibitem[{OpenAI et~al.(2024{\natexlab{b}})OpenAI, Achiam, Adler, Agarwal, Ahmad et~al.}]{openai2024gpt4technicalreport}
OpenAI; Achiam, J.; Adler, S.; Agarwal, S.; Ahmad, L.; et~al. 2024{\natexlab{b}}.
\newblock GPT-4 Technical Report.
\newblock arXiv:2303.08774.

\bibitem[{Poole et~al.(2022)Poole, Jain, Barron, and Mildenhall}]{poole2022dreamfusiontextto3dusing2d}
Poole, B.; Jain, A.; Barron, J.~T.; and Mildenhall, B. 2022.
\newblock DreamFusion: Text-to-3D using 2D Diffusion.
\newblock arXiv:2209.14988.

\bibitem[{Roller et~al.(2021)Roller, Dinan, Goyal, Ju, Williamson, Liu, Xu, Ott, Smith, Boureau, and Weston}]{roller2021recipes}
Roller, S.; Dinan, E.; Goyal, N.; Ju, D.; Williamson, M.; Liu, Y.; Xu, J.; Ott, M.; Smith, E.~M.; Boureau, Y.-L.; and Weston, J. 2021.
\newblock Recipes for building an open-domain chatbot.
\newblock In \emph{Proceedings of the 16th Conference of the European Chapter of the Association for Computational Linguistics: Main Volume}, 300--325.

\bibitem[{Rombach et~al.(2022)Rombach, Blattmann, Lorenz, Esser, and Ommer}]{rombach2022high}
Rombach, R.; Blattmann, A.; Lorenz, D.; Esser, P.; and Ommer, B. 2022.
\newblock High-Resolution Image Synthesis with Latent Diffusion Models.
\newblock In \emph{Proceedings of the IEEE/CVF Conference on Computer Vision and Pattern Recognition}, 10684--10695.

\bibitem[{Scarlatos, Baker, and Lan(2025)}]{scarlatos2025exploring}
Scarlatos, A.; Baker, R.~S.; and Lan, A. 2025.
\newblock Exploring knowledge tracing in tutor-student dialogues using llms.
\newblock In \emph{Proceedings of the 15th International Learning Analytics and Knowledge Conference}, 249--259.

\bibitem[{Searle(1969)}]{searle1969speech}
Searle, J.~R. 1969.
\newblock \emph{Speech Acts: An Essay in the Philosophy of Language}.
\newblock Cambridge University Press.

\bibitem[{See et~al.(2019)See, Roller, Kiela, and Weston}]{see2019makesgoodconversationcontrollable}
See, A.; Roller, S.; Kiela, D.; and Weston, J. 2019.
\newblock What makes a good conversation? How controllable attributes affect human judgments.
\newblock arXiv:1902.08654.

\bibitem[{Shannon(1948)}]{shannon1948mathematical}
Shannon, C.~E. 1948.
\newblock A Mathematical Theory of Communication.
\newblock \emph{The Bell System Technical Journal}, 27(3): 379--423.

\bibitem[{Shuster et~al.(2022)Shuster, Xu, Komeili, Ju et~al.}]{shuster2022blenderbot3deployedconversational}
Shuster, K.; Xu, J.; Komeili, M.; Ju, D.; et~al. 2022.
\newblock BlenderBot 3: a deployed conversational agent that continually learns to responsibly engage.
\newblock arXiv:2208.03188.

\bibitem[{Sweller, Van~Merrienboer, and Paas(2011)}]{sweller2011cognitive}
Sweller, J.; Van~Merrienboer, J. J.~G.; and Paas, F. G. W.~C. 2011.
\newblock Cognitive Load Theory, Learning Difficulty, and Instructional Design.
\newblock \emph{Learning and Instruction}, 4(4): 295--312.

\bibitem[{Traum and Larsson(2003)}]{traum2003information}
Traum, D.~R.; and Larsson, S. 2003.
\newblock The information state approach to dialogue management.
\newblock In \emph{Current and new directions in discourse and dialogue}, 325--353. Springer.

\bibitem[{Urbanek et~al.(2019)Urbanek, Fan, Karamcheti, Jain, Humeau, Dinan, Rocktäschel, Kiela, Szlam, and Weston}]{urbanek2019learning}
Urbanek, J.; Fan, A.; Karamcheti, S.; Jain, S.; Humeau, S.; Dinan, E.; Rocktäschel, T.; Kiela, D.; Szlam, A.; and Weston, J. 2019.
\newblock Learning to Speak and Act in a Fantasy Text Adventure Game.
\newblock In \emph{Proceedings of the 2019 Conference on Empirical Methods in Natural Language Processing}, 673--683.

\bibitem[{Vaidyam et~al.(2019)Vaidyam, Wisniewski, Halamka, Kashavan, and Torous}]{vaidyam2019chatbots}
Vaidyam, A.~N.; Wisniewski, H.; Halamka, J.~D.; Kashavan, M.~S.; and Torous, J.~B. 2019.
\newblock Chatbots and Conversational Agents in Mental Health: A Review of the Psychiatric Landscape.
\newblock \emph{The Canadian Journal of Psychiatry}, 64(7): 456--464.

\bibitem[{Welleck et~al.(2019)Welleck, Weston, Szlam, and Cho}]{welleck2019dialoguenaturallanguageinference}
Welleck, S.; Weston, J.; Szlam, A.; and Cho, K. 2019.
\newblock Dialogue Natural Language Inference.
\newblock arXiv:1811.00671.

\bibitem[{Williams et~al.(2016)Williams, Henderson, Raux, Thomson, Black, and Ramachandran}]{williams2016dialog}
Williams, J.~D.; Henderson, M.; Raux, A.; Thomson, B.; Black, A.; and Ramachandran, D. 2016.
\newblock The Dialog State Tracking Challenge Series: A Review.
\newblock In \emph{Dialogue \& Discourse}, volume~8, 1--33.

\bibitem[{Xu, Szlam, and Weston(2022)}]{xu2022beyond}
Xu, J.; Szlam, A.; and Weston, J. 2022.
\newblock Beyond Goldfish Memory: Long-Term Open-Domain Conversation.
\newblock In \emph{Proceedings of the 60th Annual Meeting of the Association for Computational Linguistics}, 5180--5197.

\bibitem[{Xu, Cao, and de~Polavieja(2020)}]{xu2020theory}
Xu, Y.; Cao, Z.; and de~Polavieja, G.~G. 2020.
\newblock A Theory of Usable Information Under Computational Constraints.
\newblock \emph{Entropy}, 22(9): 1014.

\bibitem[{Yang and Klein(2021)}]{Yang_2021}
Yang, K.; and Klein, D. 2021.
\newblock FUDGE: Controlled Text Generation With Future Discriminators.
\newblock In \emph{Proceedings of the 2021 Conference of the North American Chapter of the Association for Computational Linguistics: Human Language Technologies}. Association for Computational Linguistics.

\bibitem[{Young et~al.(2013)Young, Gašić, Thomson, and Williams}]{young2013pomdp}
Young, S.; Gašić, M.; Thomson, B.; and Williams, J.~D. 2013.
\newblock {POMDP}-Based Statistical Spoken Dialog Systems: A Review.
\newblock In \emph{Proceedings of the IEEE}, volume 101, 1160--1179.

\bibitem[{Zhang et~al.(2018)Zhang, Dinan, Urbanek, Szlam, Kiela, and Weston}]{zhang2018personalizing}
Zhang, S.; Dinan, E.; Urbanek, J.; Szlam, A.; Kiela, D.; and Weston, J. 2018.
\newblock Personalizing Dialogue Agents: I Have a Dog, Do You Have Pets Too?
\newblock In \emph{Proceedings of the 56th Annual Meeting of the Association for Computational Linguistics}, 2204--2213.

\bibitem[{Zhang et~al.(2020)Zhang, Sun, Galley, Chen, Brockett, Gao, Gao, Liu, and Dolan}]{zhang2020dialogptlargescalegenerativepretraining}
Zhang, Y.; Sun, S.; Galley, M.; Chen, Y.-C.; Brockett, C.; Gao, X.; Gao, J.; Liu, J.; and Dolan, B. 2020.
\newblock DialoGPT: Large-Scale Generative Pre-training for Conversational Response Generation.
\newblock arXiv:1911.00536.

\bibitem[{Zheng et~al.(2023)Zheng, Chiang, Sheng, Zhuang, Wu, Zhuang, Lin, Li, Li, Xing et~al.}]{zheng2023judging}
Zheng, L.; Chiang, W.-L.; Sheng, Y.; Zhuang, S.; Wu, Z.; Zhuang, Y.; Lin, Z.; Li, Z.; Li, D.; Xing, E.; et~al. 2023.
\newblock Judging llm-as-a-judge with mt-bench and chatbot arena.
\newblock \emph{Advances in neural information processing systems}, 36: 46595--46623.

\end{thebibliography}

\appendix
\setcounter{secnumdepth}{1}


\section{Prompts}
In this section, we present the prompt used for conversation generation.
\subsection{Raw Prompt}
Create a K-turn conversation between an AI adviser and an entrepreneur trying to work on $\textless$A business field$\textgreater$. In the conversation, the AI adviser is an informed business coach in a Small Business Development Corporation, and the entrepreneur is a $\textless$ entrepreneur's demographic background $\textgreater$ with a focus on $\textless$entrepreneur's idea$\textgreater$.
\subsection{Parameterized Prompt}
\label{subsec:sim_parameters}

Below is the complete prompt to the LLM for parameterized conversation generation:

\subsection{Conversation Parameters Structure}
\label{subsec:param_structure}

The conversation generator operates using a hierarchical parameter system organized into six main categories: Fundamentals, Participants, Learning Approach, Conversation Dynamics, Linguistic Patterns, and Content Attributes.

\subsection{Fundamentals}
\label{subsec:fundamentals}

Core structural parameters that define the conversation's basic framework:

\begin{itemize}
    \item \textbf{Purpose:} The primary intent of the conversation
    \begin{itemize}
        \item \textit{advisory:} Problem-solving and guidance-focused dialogue
        \item \textit{educational:} Knowledge transfer and learning-oriented
        \item \textit{exploratory:} Discovery and brainstorming-centered
        \item \textit{evaluative:} Assessment and critique-focused
    \end{itemize}
    
    \item \textbf{Turns:} Total number of conversation turns (exchanges between participants)
    
    \item \textbf{Turn Balance:} Distribution of conversation contributions between participants (expressed as ratio, e.g., "55:45" means user speaks 55\% of turns, advisor 45\%)
    
    \item \textbf{Arc:} Overall narrative structure of the conversation
    \begin{itemize}
        \item \textit{problem-solution:} Identifies issues and develops solutions
        \item \textit{exploration-conclusion:} Broad investigation leading to specific outcomes
        \item \textit{question-answer:} Sequential inquiry and response pattern
        \item \textit{build-refine:} Iterative development and improvement process
    \end{itemize}
    
    \item \textbf{Initiator:} Which participant starts the conversation
    \begin{itemize}
        \item \textit{user:} Entrepreneur begins with question or problem
        \item \textit{assistant:} Advisor opens with inquiry or observation
    \end{itemize}
    
    \item \textbf{Topic Scope:} Array of subject areas that may be covered during the conversation (e.g., ["food business", "marketing", "operations"])
\end{itemize}

\subsection{Participants}
\label{subsec:participants}

Parameters defining the characteristics and relationship between conversation participants:

\begin{itemize}
    \item \textbf{Knowledge Gap Level (KGL)}
\begin{itemize}
    \item \textbf{1:} Expert with deep understanding of business domain
    \item \textbf{2:} Advanced practitioner with solid foundational knowledge and some specialized expertise
    \item \textbf{3:} Moderate familiarity with business concepts
    \item \textbf{4:} Basic understanding with significant knowledge gaps requiring guidance
    \item \textbf{5:} Complete novice with minimal business knowledge about their ideas
\end{itemize}
    
    \item \textbf{Assistant Parameters:}
    \begin{itemize}
        \item \textbf{Identity:} Role and background description (e.g., "experienced business advisor with small business expertise")
        \item \textbf{Consistency Level:} How consistently the assistant maintains their role and expertise (0.0 = highly variable, 1.0 = perfectly consistent)
    \end{itemize}
    
    \item \textbf{User Parameters:}
    \begin{itemize}
        \item \textbf{Identity:} Role and background description (e.g., "early-stage food business entrepreneur")
        \item \textbf{Focus Level (FL)}
            \begin{itemize}
                \item \textbf{1:} Free-flowing, wide-ranging conversation covering many aspects
                \item \textbf{2:} Mostly broad discussion with occasional deep dives into specific areas
                \item \textbf{3:} Balanced focus with some exploration of tangential topics
                \item \textbf{4:} Primarily focused on core issues with minimal tangential exploration
                \item \textbf{5:} Laser-focused on specific details of implementation
            \end{itemize}
        \item \textbf{Prior Knowledge Level:} User's existing expertise in the domain (1 = complete novice, 2 = limited knowledge, 3 = moderate level understanding, 4 = extensive previous experience, 5 = expert level)
        \item \textbf{Decision-Making Style (DMS)}
            \begin{itemize}
                \item \textbf{Analytical:} Focuses on data, metrics, and logical analysis
                \item \textbf{Intuitive:} Relies on gut feeling and personal judgment
                \item \textbf{Consultative:} Seeks multiple perspectives before deciding
                \item \textbf{Risk-averse:} Prioritizes minimizing potential downsides
                \item \textbf{Impulsive:} Makes quick decisions without extensive deliberation
            \end{itemize}
        \item \textbf{Feedback Reception (FR)}
        \begin{itemize}
            \item \textbf{Receptive:} Eagerly accepts and builds upon advice
            \item \textbf{Balanced:} Considers advice thoughtfully with moderate acceptance
            \item \textbf{Skeptical:} Questions most suggestions, needs convincing
            \item \textbf{Resistant:} Pushes back against most advice, difficult to persuade
        \end{itemize}
    \end{itemize}
\end{itemize}

\subsection{Learning Approach}
\label{subsec:learning_approach}

Parameters controlling how knowledge is delivered and educational objectives are achieved:

\begin{itemize}
    \item \textbf{Framework:} Educational methodology employed
    \begin{itemize}
        \item \textit{socratic:} Question-driven discovery learning
        \item \textit{didactic:} Direct instruction and explanation
        \item \textit{collaborative:} Joint problem-solving approach
        \item \textit{experiential:} Learning through practical examples and scenarios
    \end{itemize}
    
    \item \textbf{Practical-Theoretical Balance:} Ratio of practical application to theoretical concepts (0.0 = purely theoretical, 1.0 = purely practical)
    
    \item \textbf{Complexity Progression:} Array showing how conceptual difficulty increases throughout the conversation (e.g., [0.3, 0.5, 0.7, 0.8] indicates gradual complexity increase)
    
    \item \textbf{Industry Context:} Specific sector or domain focus (e.g., "food-business", "technology", "healthcare")
\end{itemize}

\subsection{Conversation Dynamics}
\label{subsec:conversation_dynamics}

Parameters governing interpersonal interactions and emotional progression:

\begin{itemize}
    \item \textbf{Formality:} Level of professional versus casual communication (0.0 = highly casual, 1.0 = highly formal)
    
    \item \textbf{Emotional Journey:} Array of emotional states and their intensities throughout the conversation
    \begin{itemize}
        \item Each entry contains an emotion and intensity level (0.0 = minimal, 1.0 = maximum)
        \item Example: [{"uncertainty": 0.8}, {"curiosity": 0.7}, {"confusion": 0.5}, {"understanding": 0.6}, {"confidence": 0.7}]
    \end{itemize}
    
    \item \textbf{Relationship Development:} How much the participant relationship evolves during the conversation (0.0 = static relationship, 1.0 = significant relationship building)
    
    \item \textbf{Disagreement Handling:} Approach to managing conflicting viewpoints
    \begin{itemize}
        \item \textit{diplomatic:} Respectful acknowledgment and gentle correction
        \item \textit{direct:} Clear, straightforward disagreement
        \item \textit{avoidant:} Minimizing or redirecting conflict
        \item \textit{collaborative:} Working together to resolve differences
    \end{itemize}
\end{itemize}

\subsection{Linguistic Patterns}
\label{subsec:linguistic_patterns}

Parameters controlling language use and communication style:

\begin{itemize}
    \item \textbf{Technical Language Level:} Degree of specialized terminology and jargon (0.0 = plain language only, 1.0 = highly technical)
    
    \item \textbf{Question Types:} Distribution of different inquiry styles
    \begin{itemize}
        \item \textbf{Closed:} Yes/no or specific factual questions
        \item \textbf{Open:} Broad, exploratory questions requiring detailed responses
        \item \textbf{Rhetorical:} Questions posed for emphasis rather than response
        \item \textbf{Clarifying:} Questions seeking to understand or confirm information
        \item Values should sum to 1.0 (e.g., {"closed": 0.2, "open": 0.5, "rhetorical": 0.1, "clarifying": 0.2})
    \end{itemize}
    
    \item \textbf{Response Style:} Communication characteristics
    \begin{itemize}
        \item \textbf{Conciseness:} Brevity versus elaboration (0.0 = very verbose, 1.0 = extremely concise)
        \item \textbf{Directness:} Straightforward versus indirect communication (0.0 = highly indirect, 1.0 = completely direct)
        \item \textbf{Formality:} Professional versus casual language (0.0 = very casual, 1.0 = highly formal)
    \end{itemize}
\end{itemize}

\subsection{Content Attributes}
\label{subsec:content_attributes}

Parameters ensuring quality and comprehensiveness of conversation content:

\begin{itemize}
    \item \textbf{Factual Accuracy:} Degree of correctness in information provided (0.0 = potentially inaccurate, 1.0 = verified accuracy)
    
    \item \textbf{Example Specificity:} Level of detail in illustrations and case studies (0.0 = general examples, 1.0 = highly specific, detailed examples)
    
    \item \textbf{Stakeholder Perspectives:} Array of viewpoints to be considered during the conversation (e.g., ["customer", "supplier", "regulator", "competitor"])
\end{itemize}

\subsection{Implementation Guidelines}
\label{subsec:implementation}

When generating conversations using these parameters:

\begin{enumerate}
    \item Begin by establishing participant identities and knowledge levels
    \item Follow the specified conversation arc while maintaining turn balance
    \item Progress complexity according to the defined progression array
    \item Incorporate emotional journey elements at appropriate conversation points
    \item Ensure content addresses multiple stakeholder perspectives
    \item Maintain consistency with linguistic pattern specifications
    \item Adapt technical language level to participant knowledge asymmetry
\end{enumerate}

\subsection{Parameter Validation}
\label{subsec:validation}

Before conversation generation, validate that:
\begin{itemize}
    \item All numerical parameters fall within specified ranges (0.0-1.0)
    \item Question type distributions sum to 1.0
    \item Turn balance ratios are mathematically consistent
    \item Complexity progression shows logical advancement
    \item Stakeholder perspectives are relevant to industry context
\end{itemize}

\subsection{Output Format}
\label{subsec:output_format}

Generated conversations should follow this structure:

\begin{verbatim}
{
  "metadata": {
    "participantRoles": {...},
    "conversationArc": "...",
    "totalTurns": n
  },
  "conversation": [
    {
      "turn": 1,
      "speaker": "user|assistant",
      "content": "...",
      "emotionalState": "...",
      "complexityLevel": 0.x
    },
    ...
  ],
  "analysis": {
    "parameterAdherence": {...},
    "learningObjectivesMet": [...],
    "stakeholderPerspectivesCovered": [...]
  }
}
\end{verbatim}

Here is an example input about a user's background:
\begin{verbatim}
{
  "conversationParameters": {
    "fundamentals": {
      "purpose": "advisory",
      "turns": 12,
      "turnBalance": "55:45",
      "arc": "problem-solution",
      "initiator": "user",
      "topicScope": 
           ["food business",
           "marketing", "operations"]
    },
    "participants": {
      "knowledgeGapLevel": 3,
      "assistant": {
        "identity": 
            "experienced business advisor",
        "consistencyLevel": 0.85
      },
      "user": {
        "identity": 
            "early-stage food"
            "business entrepreneur",
        "focusLevel": 3,
        "priorKnowledgeLevel": 0.4,
        "decisionMakingStyle": "analytical",
        "feedbackReception": "receptive"
      }
    },
    "learningApproach": {
      "framework": "socratic",
      "practicalTheoreticalBalance": 0.7,
      "complexityProgression": 
          [0.3, 0.5, 0.7, 0.8],
      "industryContext": "food-business"
    },
    "conversationDynamics": {
      "formality": 0.7,
      "emotionalJourney": [
        {"uncertainty": 0.8},
        {"curiosity": 0.7},
        {"understanding": 0.6},
        {"confidence": 0.7}
      ],
      "relationshipDevelopment": 0.5,
      "disagreementHandling": "diplomatic"
    },
    "linguisticPatterns": {
      "technicalLanguageLevel": 0.6,
      "questionTypes": {
        "closed": 0.2,
        "open": 0.5,
        "rhetorical": 0.1,
        "clarifying": 0.2
      },
      "responseStyle": {
        "conciseness": 0.5,
        "directness": 0.6,
        "formality": 0.7
      }
    },
    "contentAttributes": {
      "factualAccuracy": 0.9,
      "exampleSpecificity": 0.6,
      "stakeholderPerspectives": 
      ["customer", "supplier", 
      "regulator", "competitor"]
    }
  }
}
\end{verbatim}
\section{More Results}

\subsection{Baseline Performance Comparison}
We compare the performance of different models in terms of topic diversity and topic entropy when given the baseline prompt. (Table \ref{tab:topic-div-entropy}). The result shows \textit{claude-3.7-sonnet} has the best topic diversity, and smaller models like \textit{llama3.1:70b} have the same poor performance compared to the parameterized version. 
\begin{table}[t]
  \centering
  \begin{tabular}{lrr}
    \hline
    Model         & Topic diversity & Topic entropy \\
    \hline
    claude        & 25 & 2.366 \\
    deepseek‑r1   & 18 & 2.195 \\
    o3            & 27 & 2.493 \\
    o4‑mini       & 33 & 2.880 \\
    gpt‑4.1       & 31 & 2.762 \\
    gpt‑4o‑mini   &  12 & 1.012 \\
    gemini-2.5-pro        & 28 & 2.511 \\
    llama3.1:70b        &   5 & 0.810 \\
    claude-3.7-sonnet      &  35 & 2.985 \\
    \hline
  \end{tabular}
  \caption{Topic diversity and topic entropy of baseline models.}
  \label{tab:topic-div-entropy}
\end{table}

\subsection{More Parameter Adherence Results}
\paragraph{Experience Level}
We categorize the experience level using the prior knowledge level in the original prompt and calculate the MSE between the actual and predicted value. All models show a decrease in MSE with higher turns. (Figure \ref{fig:revisit_vertical_tabular})

\paragraph{Feedback Reception}
The measurement of feedback reception is categorized into four types described in the prompt, and the result is calculated based on the rate of correct classification. The response indicates that some advanced models achieve a very high level of accuracy by combining a mixture of LLM and human decision-making, demonstrating that these models can accurately simulate the user's sentiment based on a description. Other advanced models and small models show less optimal results in this role-playing setting. (Figure \ref{fig:revisit_vertical_tabular})

\begin{figure}[t]
  \centering
  \setlength{\tabcolsep}{0pt}
  \begin{tabular}{@{}c@{}}
    \includegraphics[width=0.8\columnwidth]{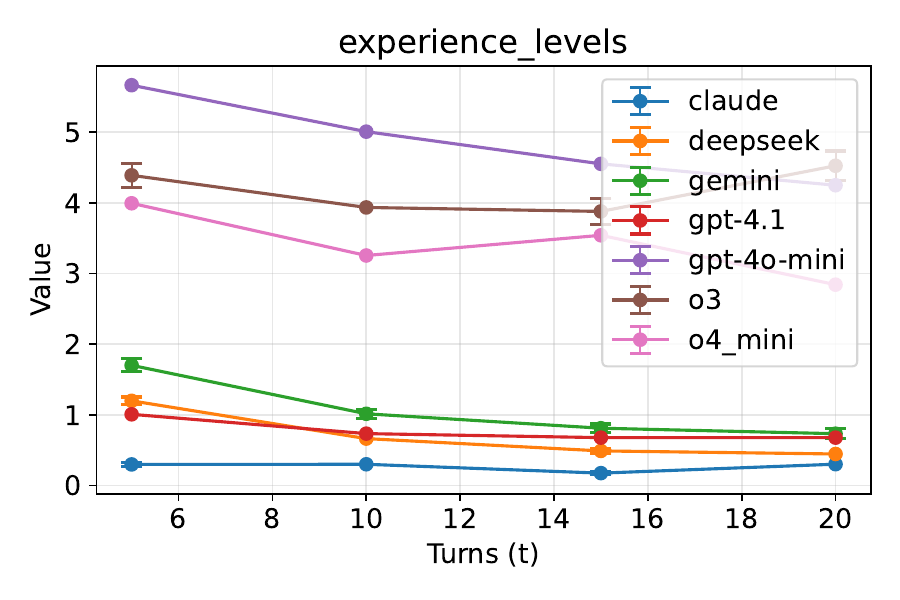} \\
    \includegraphics[width=0.8\columnwidth]{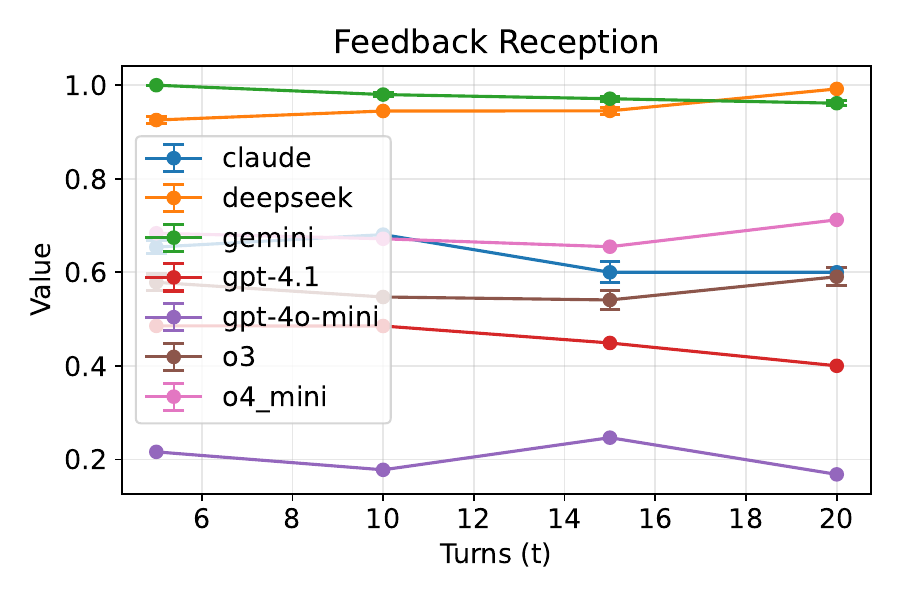} \\
  \end{tabular}
  \caption{Additional figures on parameter adherence}
  \label{fig:revisit_vertical_tabular}
\end{figure}

\end{document}